\documentclass[final,5p,times,twocolumn]{elsarticle}

\usepackage{amssymb}
\usepackage{diagbox}
\usepackage{tabularx}
\usepackage{multirow}
\usepackage{array}
\usepackage{mathtools,booktabs}
\usepackage{adjustbox}
\usepackage{makecell}
\usepackage{algorithm}
\usepackage{algpseudocode}
\usepackage{graphicx}
\usepackage{indentfirst}
\usepackage{float}
\usepackage{placeins}
\usepackage{fancyhdr}
\usepackage{multicol}
\usepackage[labelfont=bf]{caption}
\usepackage{xcolor}
\usepackage{url}
\usepackage[switch]{lineno}
\usepackage[colorlinks]{hyperref}

\AtBeginDocument{
\hypersetup{
  colorlinks=true,
  linkcolor=cyan,
  filecolor=cyan,
  citecolor=cyan,
  urlcolor=cyan
}}
 
\begin{document}
\begin{frontmatter}


\title{ Applying Self-supervised Learning to Network Intrusion Detection for Network Flows with Graph Neural Network }


\author[csuft]{Renjie Xu}
\ead{renjiexu@csuft.edu.cn}

\author[csuft]{Guangwei Wu\corref{mycorrespondingauthor}}
\cortext[mycorrespondingauthor]{Corresponding author}
\ead{guangweiwu@csuft.edu.cn}

\author[csu]{Weiping Wang}
\ead{wpwang@csu.edu.cn}

\author[csuft,csu]{Xing Gao}
\ead{20202756@csuft.edu.cn}

\author[csuft]{An He}
\ead{an_he@csuft.edu.cn}

\author[csuft]{Zhengpeng Zhang}
\ead{zhengpeng@csuft.edu.cn}

\affiliation[csuft]{
    organization={College of Computer and Information Engineering, Central South University of Forestry and Technology},
    city={Changsha},
    postcode={410004}, 
    country={P.R.~China}       
}
\affiliation[csu]{
    organization={School of Computer Science and Engineering, Central South University},
    city={Changsha},
    postcode={410083}, 
    country={P.R.~China}       
}

\begin{abstract}
Graph Neural Networks (GNNs) have garnered intensive attention for Network Intrusion Detection System (NIDS) due to their suitability for representing the network traffic flows.
However, most present GNN-based methods for NIDS are supervised or semi-supervised. 
Network flows need to be manually annotated as supervisory labels, a process that is time-consuming or even impossible, making NIDS difficult to adapt to potentially complex attacks, especially in large-scale real-world scenarios.
The existing GNN-based self-supervised methods focus on the binary classification of network flow as benign or not, and thus fail to reveal the types of attack in practice. 
This paper studies the application of GNNs to identify the specific types of network flows in an unsupervised manner.
We first design an encoder to obtain graph embedding, that introduces the graph attention mechanism and considers the edge information as the only essential factor. 
Then, a self-supervised method based on graph contrastive learning is proposed. 
The method samples center nodes, and for each center node, generates subgraph by it and its direct neighbor nodes, and corresponding contrastive subgraph from the interpolated graph, and finally constructs positive and negative samples from subgraphs.
Furthermore, a structured contrastive loss function based on edge features and graph local topology is introduced.
To the best of our knowledge, it is the first GNN-based self-supervised method for the multiclass classification of network flows in NIDS.
Detailed experiments conducted on four real-world databases (NF-Bot-IoT, NF-Bot-IoT-v2, NF-CSE-CIC-IDS2018, and NF-CSE-CIC-IDS2018-v2) systematically compare our model with the state-of-the-art supervised and self-supervised models, illustrating the considerable potential of our method.
Our code is accessible through \url{https://github.com/renj-xu/NEGSC}.
\end{abstract}


\end{frontmatter}


\section{Introduction}

With the exponential growth in user hosts and network services, the frequency and the complexity of cyberattacks are increasing \cite{J. Piet, Y. Qin}. 
There has been a rise in innovative cyber-attacks on vital infrastructure of Internet of Things (IoT) networks, traditional Internet environments and research institutions \cite{A.M. Mandalari, H. Jmila}. 
Protecting network security has become more challenging than ever. 
In the real world, networks are generally heterogeneous, where hundreds of millions of devices come from different manufacturers, and security protocols are difficult to align. 
The emergence of intelligent networking makes networks even more complex. 
These trends dramatically increase cybersecurity risks in networks, and pose a substantial challenge to cyber-attack detection technology.


In the highly liberal digital space, network devices are vulnerable to various cyberattacks, including data theft, denial of service (DoS), distributed denial of service (DDoS), reconnaissance attacks and brute force attack, and so on \cite{S.I. Popoola}. 
These attacks threaten the normal operation of the device and the network security. 
Indeed, the individuals may suffer from property loss and privacy breaches, and companies may be at risk of intellectual property of high economic interest, customer data or disrupt normal operations. 
Remediation is costly or even irretrievable if the attacks cannot be accurately detected. 

Network Intrusion Detection Systems (NIDSs) have proven their reliability in detecting and mitigating cyber-attacks, whose main tasks are to capture malicious network traffic flow and apply the identification outputs to achieve precise and prompt responses. 
To achieve this task, they require high-quality network flow datasets and data mining to identify potential intrusions. 
Nowadays, NetFlow is one of the main formats in NIDS for collecting and recording network traffic flows, which provides rich IP flow data and becomes an essential tool for monitoring and recording network flow in network security \cite{dataset1}. 
The flow data can effectively help us identify and locate anomalous behaviors in the network, and detect and prevent network attacks in a timely manner \cite{F. Yang}, which is of the importance in network security. 
As a consequence, NIDSs are regularly deployed at the edge of networks and crucial facilities to monitor and analyse the inbound and outbound network traffic \cite{S. Zhan}.

NIDSs can be broadly classified into two main types: Signature-based Network Intrusion Detection (SNIDS) and Behavior-based Network Intrusion Detection (BNIDS).
SNIDS apply predefined rules or signatures to detect known attack patterns, based on behavioral patterns or specific attack signatures, which trigger system alerts when there are network flows or system activities matching with the rules \cite{A.H. Qureshi}. 
SNIDS are unable to flag newly identified and unknown suspicious behaviors \cite{H. Holm}. 
Through monitoring and analyzing diverse behaviors and activities within computer systems and networks, BNIDS detect potential intrusions or threats by identifying newly found or unknown patterns of behavior. 
Their operations are now more efficient and effective thanks to the development of advanced machine learning (ML) algorithms. 
However, NIDS continue to encounter numerous obstacles due to the increasing frequency and complexity of attacks.

Extracting features from raw data of network flow samples is vital in the learning process. 
In traditional ML-based intrusion detection systems, hand-crafted data features are often designed \cite{G. Engelen}. 
Such an approach requires suitable functions or statistical metrics to extract the desired information from network flow, placing demands on human experts experience. 
Dealing with modelling cyber-attacks and building such large datasets can be expensive and time-consuming. 
Furthermore, these hand-crafted functional representations have relatively low capabilities and are hard to express deep semantic information. The massive increase in connected devices and more complex data interactions impose new demands on automated feature extraction. 
Deep learning (DL) technology can automatically extract high-level feature representations from raw data of network flows through a well-designed multi-layer neural network, thus eliminating manual feature engineering. 
DL has made significant progress in NIDS \cite{H. Jmila}. 
Through modelling network traffic detection task as a binary versus multiclass classification problem, it enables accurate identification and classification of malicious attacks.
Recently, many researchers focus on the DL methods that are capable of dealing with graphical structure data, which is the main form of network traffic flows.

\begin{figure*}[h]
    \centering
    \includegraphics[width=0.5\textwidth]{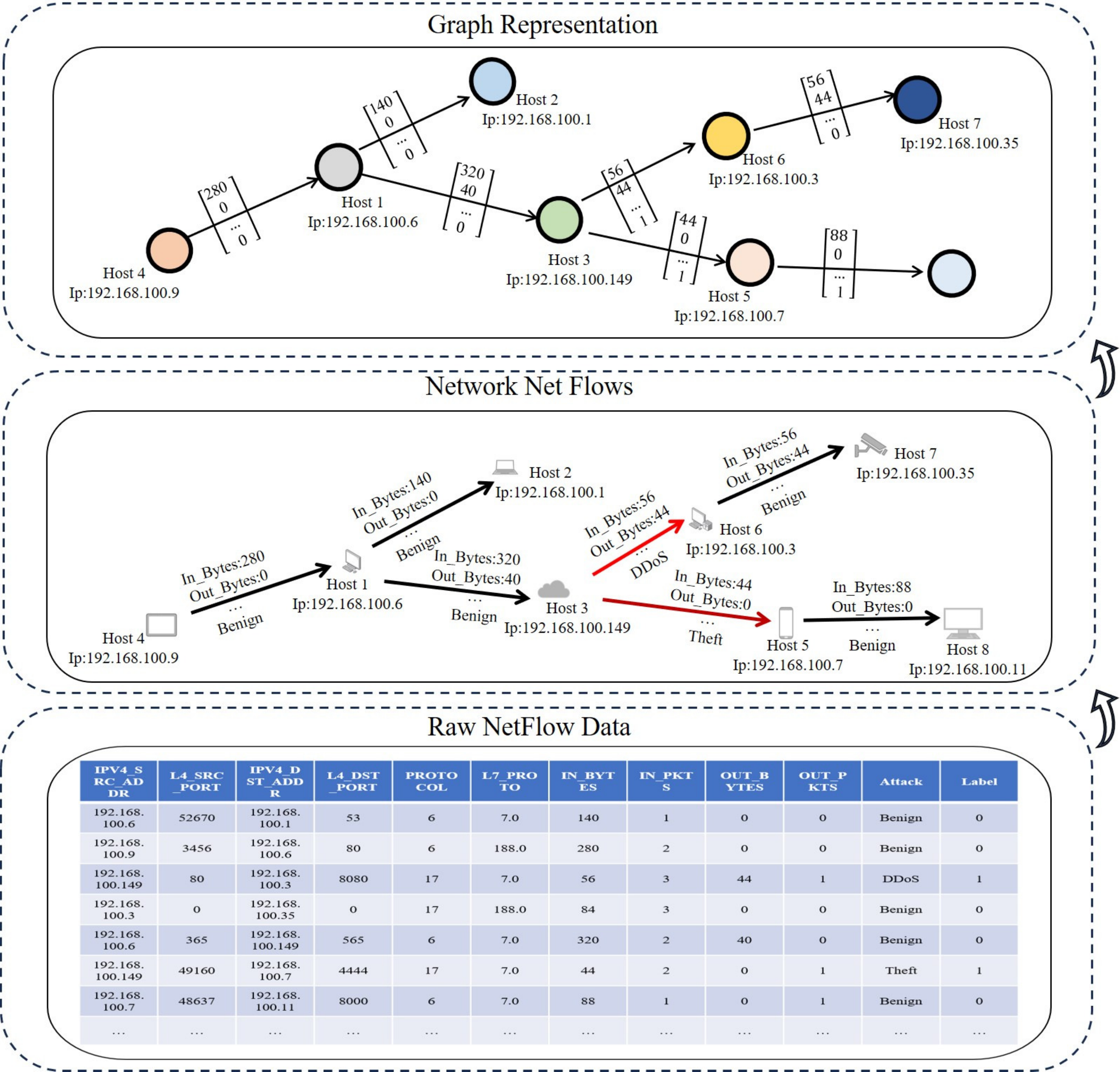}
    \caption{Converting Netflow-based data into graph representation. 
    An arrow along with nodes indicates a network traffic flow from the source host to the destination host. 
    Normal and attack flows are denoted by black and red arrows, respectively, where different shades of red arrows indicate different types of attacks.}
    \label{data to graph}
\end{figure*}

Graphs have a powerful ability to represent non-Euclidean data, and have a well-established theoretical foundation. 
Intuitively, network flows data naturally form a graph, where hosts and network flows between hosts in network can be viewed as nodes and edges between nodes in graph, respectively (See Fig. \ref{data to graph} for more details). 
An earlier research conducted by Staniford-Chen et al \cite{S. Staniford-Chen} has proposed the application of graph theory in modelling network behaviors to improve the effectiveness of NIDS in managing the collaborative attack capability of large networks. 
In recent studies \cite{I.A. Chikwendu}, the Graph Neural Networks (GNNs), which are a powerful deep learning method for graph structured data, have been introduced into NIDS.
Most DL-based NIDS directly apply GNN models \cite{W.W. Lo, H. Nguyen}. 
These models are node-centric, i.e., they rely heavily on node features but do not adequately utilize edge features. 
Node-centric representations assume that nodes are potentially related to their neighbors. 
It is not entirely suitable for extracting the interaction information of the graph constructed by network traffic flows, where the core information of the graph is concentrated in edges rather than nodes in the graph. 
Some NIDS models extend the graph neural networks to more accurately uncover anomalous behavioral characteristics behind network flow \cite{W. Hamilton}.
Specifically, these revised models accept both node and edge features and integrate edge information into the hierarchical feature representation.
But they are usually relatively straightforward, and do not fully take the advantages of GNN and the properties of network traffic flows.

Most deep learning-based NIDSs are supervised paradigm, which require numerous accurately labelled data samples \cite{F. Wei}.
These methods use supervised signals labelled by hand to establish mapping between data features and categories.
However, in practice, only a fews part of network traffic can typically be labelled.
With the varying and advancing patterns and strategies of cyber-attacks, it is unfeasible to encompass all possible forms of attacks even with a vast pool of labelled data. This requires network intrusion detection systems to be adaptive and flexible enough to effectively detect new and unknown attacks \cite{P. Rieger}. 
Labelled information is user-friendly for humans to comprehend and make decisions, but its assistance to models in generalizing knowledge may be limited, imprecise and even fragile. 
Because the intrinsic associations of the data may be not exactly precise to be simply divided by the labels, and imbalances in the fixed label categories of the samples may affect the model preferences, especially in networks with diverse attacks and noisy. 
Moreover, in real-time network environments, obtaining high-quality labels usually implies high costs or is not accessible. 
Duan et al. \cite{G. Duan} enhance the information aggregation capability of the convolution process by using a linear structured graph neural network and introduce a semi-supervised training method to reduce the reliance on labels. 
The best current anomaly detection accuracy was achieved on multiple datasets, however, fully unsupervised learning remains to be achieved.

To address the aforementioned problems, a GNN-based self-supervised representation learning is proposed for NIDS, where our focus is on edge features, i.e., network flow information and graph topology. 
Compared to conventional DL-based intrusion detection approaches, our model performs self-supervised learning that automatically reveals the underlying structure of the data without the need for manual labeling or any prior knowledge. 
This reduces the labor and time costs while expanding the scope of intrusion detection for unknown attacks in networks. 
The main contributions of this paper are as follows:

1) We propose NetFlow-Edge Generative Subgraph Contrast (NEGSC), a self-supervised graph representation learning method for identifying malicious attacks and their specific types. 
The framework deeply incorporates the features of network flows in a self-supervised graph contrast learning framework Generative Subgraph Contrast (GSC), and pays attention to the local structure of graph, which generates contrastive subgraphs by the generation module based on the center nodes and their direct neighbor nodes. This is beneficial for revealing different types of malicious traffics more elaborately.
Furthermore, a structured contrastive loss function is used to distinguish the contrastive subgraph samples by fully exploiting the local network flow structure and attributes. 

2) To efficiently utilize the raw data of network flows for constructing graph embedding fed to NEGSC, we introduce an edge-featured self-attention mechanisms, named NetFlow-Edge Graph Attention Network (NEGAT), as the encoder of NEGSC.
It makes better use of the information among adjacent network flows by weighted aggregation. 
Additionally, in NEGAT, the node acts as an auxiliary relay of network flow to obtain the encoding of adjacent edge features, thus avoiding significant computational cost for the attention weights.

Extensive experiments are conducted on four well-known modern intrusion detection datasets in both binary and multiclass classification scenarios. 
The datasets (NF-Bot-IoT, NF-Bot-IoT-v2, NF-CSE-CIC-IDS2018 and NF-CSE-CIC-IDS2018-v2) are all NetFlow-based.
In binary classification scenario, our method significantly outperforms the existing self-supervised learning method in terms of Acc, Precision, Recall and F1.
Due to that there is no unsupervised GNN-based model for multiclass classification, our method is compared to other latest  supervised GNN-based model.
The experiment indicates that our method, in a self-supervised manner, can match the supervised GNN-based model on the metrics Recall and F1.


The rest of the paper is organized as follows. 
Section \ref{2} reviews the related works. 
The proposed methods in detail are introduced in section \ref{3}. 
Section \ref{4} describes the experimental setup and results. 
In Section \ref{5}, we summarize our work and discuss the future research directions.

\section{Related Work}
\label{2}
The section mainly explores the trends of DL-based methods in NIDSs. 
These methods provide new perspectives and tools to address the increasing frequency and complexity of cyberattacks, and have profoundly impacted cybersecurity.




\subsection{DL-based NIDSs}
DL methods exhibit superior capabilities in handling large-scale data when compared with traditional ML methods. 
It autonomously extracts feature representations from raw data, eliminating the need for manual feature selection and thereby enhancing the model's expressiveness \cite{C. Janiesch}.

Wang et al. \cite{W. Wang} proposed a CNN-based NIDS, training a Deep Learning-based detection model using extracted features and raw network traffic. 
The study demonstrated superior accuracy when the model was trained on raw traffic compared to using extracted features.
Gupta et al. \cite{N. Gupta} proposed an anomaly based network intrusion detection system LIO-IDS that uses Long Short Term Memory classifiers and improved One-vs-One technique to handle network intrusions.
Jiang et al. \cite{K. Jiang} proposed a model that combines CNN and Bidirectional Long Short-Term Memory (BiLSTM) for Intrusion Detection Systems (IDS).
Le et al. \cite{T.T.H. Le} proposed a model applying the RNN and its variants, and the detection error rate of their model is lower than the baseline on NSL-KDD and ISCX data.

Although DL-based NIDSs have achieved remarkable results especially in large-scale data scenario, these methods often overlook or are incapable of dealing with the inherent graph structure of network traffic flows, that is beneficial for identifying anomalous nodes and behaviors \cite{ref52}.

\subsection{GNN-based NIDSs}
GNNs are a powerful DL method for graph-structured data.
In recent years, GNNs have attracted much attention for their excellent performance in NIDS. 
The inherent graph formation of network flows makes GNNs particularly effective in this context.
The main challenge of applying GNNs to NIDS is that the datasets in NIDS are usually edge-centered, that is, the core information is concentrated on edges (network flows) and the classification task is for edges (network flows).

Zhou et al. \cite{J. Zhou Z. Xu} proposed a GNN-based approach for automatic learning and detection of botnet strategies. 
Their end-to-end data-driven approach utilizes GNNs to capture specific botnet topologies, relying solely on graph structures for detection. 
Xiao et al. \cite{Q. Xiao} gave a network anomaly detection method for automatically learning implicit features of network traffic. 
They transformed the network traffic into first- and second-order graphs, obtained low-dimensional vector representations of nodes, and trained classifiers for anomaly detection. 
Lo et al. \cite{W.W. Lo} proposed an innovative GNN-based NIDS to detect malicious traffic in IoT networks. 
The approach, E-GraphSAGE, is an extension of the original GraphSAGE algorithm \cite{W. Hamilton}, that generates edge embeddings and implements edge classification to categorize network flows using edge features and topology information. 
These results further prove the potential and advantages of GNN in detecting network intrusion.

The aforementioned methods employ GNNs for network anomaly detection under a supervised learning manner, relying on pre-labeled data during training. 
However, labeled data represents only a tiny fraction of the total data under realistic network environment. 
On the other hand, self-supervised learning has shown great potential as an effective strategy to address the challenges of scarcity of labeled data, such as the limited availability of training data, domain-specific requirements, and vulnerability to labeling against attacks \cite{Y. Liu,Y. Xie}. 

Lo et al. \cite{E. Caville} proposed a self-supervised GNN-based approach called Anomal-E for distinguishing between normal and malicious network traffics.
Anomal-E employs E-GraphSAGE as an encoder to generate graph embedding from raw NetFlow data, and then feeds the embedding to a revised version of a self-supervised learning framework Deep Graph Infomax (DGI) \cite{P. Veličković}.
The revised version leverages edge features and graph topological structure in original DGI framework. 
The positive and negative embeddings by Anomal-E are utilized to tune and optimize E-GraphSAGE, which is used in downstream anomaly detection algorithms.
This is the first successful approach to network intrusion detection using a self-supervised GNN.
Very recently, Nguyen et al. \cite{H. Nguyen} proposed an intrusion detection method named TS-IDS with a self-supervised module. 
It converts the problem of edge classification in graph into predicting node attributes. 
The loss function is obtained by combining the self-supervised loss of the nodes and the supervised loss of the edges. 
Note that TS-IDS is still supervised, where the supervised loss is computed using edge labels, and the self-supervised module is used to enhance the graph representation.

Even though GNN-based NIDSs have received increasing interests, there is a lack of the self-supervised GNNs classification results for the problem. 
To the best of our knowledge, no GNN-based algorithm is capable for the multiclass classification task in NIDS under a self-supervised manner.
Meanwhile, many sophisticated deep learning techniques, such as attention mechanisms, have not yet been widely applied in NIDS.

\subsection{Edge-Featured Approaches}
As in NIDS, edge features play an essential role in many other fields of the real world. 
For example, in social networks, edges represent interpersonal interactions, the importance and complexity of which can be captured by information about the edges. 
Many studies have begun to focus on and utilize the properties or features of edges to enhance graph data processing.

The EGNN framework proposed by Gong and Cheng \cite{L. Gong} can more fully utilize edge features in the graph, including undirected or multidimensional edge features. 
It uses the edge features to compute the weight matrix, which assists in propagating the node features. 
Jiang et al. \cite{X. Jiang} proposed a new graph neural network framework, CensNet, which can utilize graph structure, node, and edge features to learn both node and edge embeddings. 
Chen et al. \cite{J. Chen} proposed an Edge-Featured Graph Attention Network (EGAT) that can handle both node and edge features, combining edge features and node features to compute message and attention weights so that edge features can effectively participate in learning graph feature representations.

It is worth mentioning that the above discussed methods may not be directly suitable for NIDSs.
They consider both edge data and node data as essential factors during the aggregation, and focus on node classification tasks.
But in NetFlow-based databases in NIDS, each entry (a network flow sample) represents the features of corresponding edge, and the objective is to achieve good results on edge classification tasks.



\section{Methodology}
\label{3}
In this section, we first present the notation used in the paper, and describe the data preprocessing in details. 
Then, an GNN-based unsupervised intrusion detection approach for NetFlow-based network traffic flows is introduced. 
The approach consists of two main sequential parts. 
The first part is an improved encoder named NEGAT based on GAT, which combines the attention mechanism with edges information to elaborately extract valuable features from inherent network topology and the differentiation between traffic flows. 
The second part is a self-supervised method named NEGSC based on graph contrastive learning \cite{Y. Han}, which uses NEGAT as encoder and focuses on local structure information of graph for enhancing sufficient intrusion classification performance. 
Our method achieves both binary and multiclass classification tasks for network flows via self-supervised learning.
Finally, the time complexity analysis of our model is also provided.

\subsection{Preliminary}
Denote an undirected graph $G$ by $(V, E)$, where $V$ is the set of nodes and $E$ is the set of edges in $G$. 
Let $n$ and $m$ be the number of nodes and the number of edges, respectively. 
We use $\vec{h}_v$ to represent the features of node $v$ for each $v \in V$, where $\vec{h}_v \in \mathbb{R}^{F_V}$, and $\vec{e}_{vu}$ to represent the features of edge $e_{vu}$ between nodes $v$ and $u$ for each $e_{vu} \in E$ in $G$, where $\vec{e}_{vu} \in \mathbb{R}^{F_E}$. 
$F_V$ (or $F_E$) is the dimension of node (or edge) features. 
The set $\mathcal{N}_{v}$ represents the direct neighbor nodes of node $v$.

\begin{figure*}[h]
    \centering
    \includegraphics[height=0.5\textwidth]{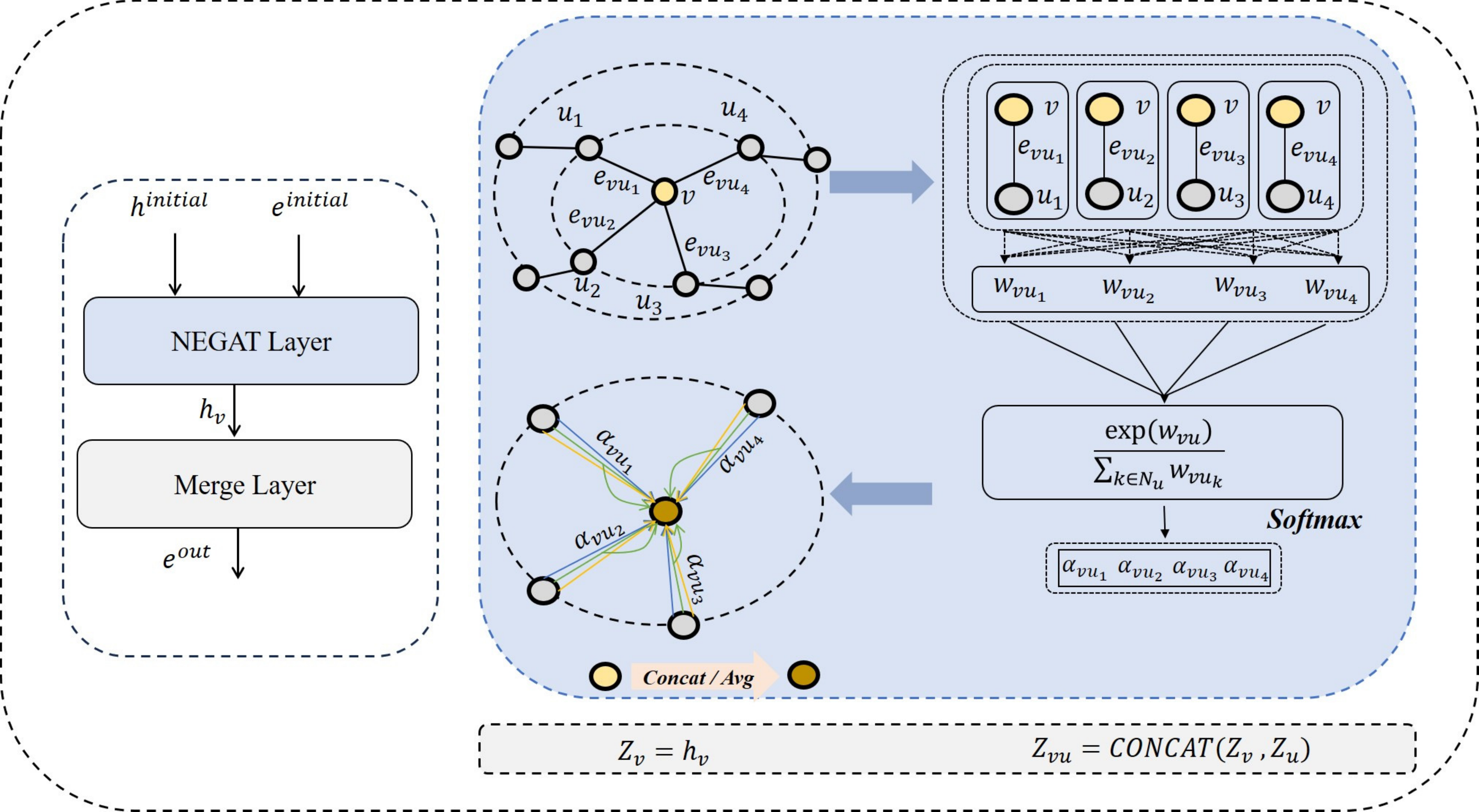}
    \caption{The architecture of NEGAT. $h$ and $e$ represent node features and edge features, respectively. The outputs of the NEGAT layer, $h_v$, are fed to the merged layer to generate the final representation of $e^{out}$. Different colored arrows indicate different attention heads, and the bottom two nodes represent the operation used in our model to concatenate attention heads.}
    \label{fig:negat}
\end{figure*}

\subsection{Data Preprocessing}

Data preprocessing plays a crucial role in transforming raw NetFlow data into graph data for our training and testing. The following provides a detailed description of our data preprocessing procedure. 
Firstly, critical information for each flow is extracted from a large amount of NetFlow data in datasets. This information includes the source IP address, destination IP address, source port number, destination port number, TCP markers, byte counts and protocol type. Forming the basis for constructing the graph data, the source IP address and destination IP address are considered as nodes, while information such as TCP markers, protocol type, and byte counts are treated as edge features. 
Secondly, direct training of all data would consume too many resources due to the large size of some original datasets. As a result, we opt for the down-sampling method (grouping datasets by ``Attack'' attribute and then taking random samples from each group) to reduce the overall data size. The down-sampled data is then divided into a training set and a testing set, with the training set used to train our model and the testing set used to evaluate its performance. Attention is paid to maintaining the consistency of the data distribution during the dataset division. 
Thirdly, the data is coded to transform categorical features into numerical features. Subsequently, normalization using the StandardScaler function is applied, ensuring that feature values are uniformly distributed within a standard normal distribution with mean 0 and variance 1. This approach helps avoid dominance of the model by specific dimensions with excessively large eigenvalues. 
Finally, the data is converted into a graph representation compatible with GNN.

\subsection{Edge-featureed Encoder}
Since in NIDS, data is usually organized by NetFlow representing traffic flows between hosts in network, critical information predominantly resides in edge features rather than node features.
Notably, in numerous GNNs, including models like GraphSAGE, the process of propagating and aggregation is primarily centered around nodes, with edges playing a supporting role. 
The risk of overlooking crucial edge information arises if attention is solely directed towards node features.  Furthermore, the advanced GNN techniques are rarely used in NIDS, such as attention mechanisms.
Despite the relative maturity of applying attention mechanisms to nodes, exploration into their application on edges remains underexplored. 
Hence, there is a need for an encoder for graph embedding, that imperatively leverages the significance of edge features to facilitate a more comprehensive capture of graph structure information, thereby enhancing the overall performance of our proposed model.

\subsubsection{ Graph Attention Network }
For a host in realistic network environment, the impacts of its related traffic flows are different.
Attention mechanism can help calculate different attention coefficients for each edge during the aggregation process, and thus should enable a more accurate capture and utilization of the importance of edges.

Velickovic et al. \cite{P. Velickovic} introduced the Graph Attention Network (GAT), which currently stands as the predominant approach for incorporating attention in GNN \cite{S. Brody}. 
It employs an attentional mechanism during the processing of input data, facilitating the automatic learning of relationships between nodes. 
Specifically, GAT assigns different weights to edges connecting a node with its neighboring nodes.
Motivated by real-world applications, an Edge-Featured Graph Attention Network (EGAT) is proposed, which is an extension of the GAT and can handle graph learning tasks with node and edge features \cite{J. Chen}. 
In EGAT, both node and edge features play important roles when computing information and attention weights. 
EGAT proposes an edge feature update mechanism that updates edge features by integrating features of connected nodes, and then uses a multi-scale merge strategy to connect the features at each level to construct a hierarchical representation.
Essentially, EGAT is still node-based, with edge information as an adjunct.

Both GAT and EGAT cannot be directly applied to NIDSs. 
In NIDSs, the critical information is on network flows between network hosts, and intrusion detection tasks also focus on network flows, that is the edges in graph.
This suggests that we further explore and develop an encoder to introduce attention mechanism to graph neural network models, that is able to use the edge information fully and accurately, and thus can better handle edge-dominated tasks in NIDSs.

\subsubsection{ The Proposed Encoder NEGAT }
Due to that the existing graph attention networks do not align with the network flow structure in NIDSs, we propose an NetFlow-Edge Graph Attention Network (NEGAT).
Compared with traditional GAT and EGAT, NEGAT focuses on extracting edge information from NetFlow-based data through attention mechanism.
It works as an edge-featured encoder that generates edge embeddings for the following self-supervised framework NEGSC.
Specifically, the proposed NEGAT first incorporates an attention mechanism into the edges to capture and utilize the importance of the edges more accurately.
Then in the process of propagating, NEGAT considers edge data as an essential factor, and node data as transit port for its adjacent edges data, thus avoiding the lift of computational cost by the aggregation of node features. 
This approach distinguishes NEGAT from existing models and is suitable for the complex network flow structure in NIDS.
The details are given in Algorithm \ref{alg:NEGAT} and Fig.\ref{fig:negat}.

\begin{algorithm}
    \caption{The NEGAT algorithm}
    \label{alg:NEGAT}
    \begin{algorithmic}[1]
    \Require {
        \Statex A graph $G = (V, E)$;  
        \Statex edge features $\{\vec{e}_{vu},\forall {e}_{vu}\in{{E}}\}$;  
        \Statex node features $\{ \vec{h}_{v}=[1,\ldots,1], \forall {v} \in {{V}} \}$; 
        \Statex Layer $K$; 
        \Statex Neighborhood function $\mathcal{N}_{v}$; 
        \Statex Weight matrix $W^{k},\forall\mathrm{k}\in\{1,\ldots, K \}$; 
        \Statex Non-linearity ${\boldsymbol{\sigma}}$;
            } 
    \Ensure {
        \Statex Embeddings $\vec{z}_{vu}; $
            }
    \State $\vec {h}_{v}^{0}\leftarrow \vec{h}_{v},\forall v\in{V}$
    \For{$k = 1$ to $K$}
        \For{$ v \in {V} $}
            \For{$u \in \mathcal{N}_{v}$}
                \State \textit {$w_{vu}^{k}=a(\boldsymbol{W^{k}}\vec{h}_{v}^{k-1} || \boldsymbol{W^{k}}\vec{e}_{vu}^{k-1} || \boldsymbol{W^{k}}\vec{h}_{u}^{k-1})$}
            \EndFor
            \State \textit {$\alpha_{vu}^{k}=softmax_{u}(w_{vu}^{k})$}
            \State \textit {$\vec {h}_{v}^{k}\longleftarrow\sigma(\sum_{u \in \mathcal{N}_{v}}\alpha_{vu}^{k}\boldsymbol{w}_{1}^{k}(\vec{h}_{v}^{k-1} || \vec{e}_{vu}^{k-1} || \vec{h}_{u}^{k-1}))$}
        \EndFor
    \EndFor
    \For{$ v\in{V} $}
        \State $  \Vec{z}_v = \Vec{h}^{K}_v$ 
    \EndFor   
    \For{$ e_{vu} \in{E} $}
        \State \textit {$\vec {z}_{vu}\leftarrow CONCAT \bigl(\vec{z}_{v},\vec{z}_{u}\bigr)$}
    \EndFor
    \end{algorithmic}
\end{algorithm}

The input of NEGAT consists of a graph $G = (V, E)$ with $n$ nodes and $m$ edges, and node feature vectors $\vec{h}_v$ for all $v \in V$ and edge feature vectors $\vec{e}_{uv}$ for all $e_{uv} \in E$. 
Note that the attributes in an $\vec{e}_{uv}$ are mapping from the properties of an entry in database, correspondingly, after preprocessing, and all $\vec{h}_v$ are initialized as $[1,\ldots,1]$ (i.e., there is no initial information on nodes).
$K$ is a hyperparameter representing the number of NEGAT layers.

Line 1 of Algorithm 1 sets the edge feature vectors as the initial embedding. 

Lines 3-9 describe an NEGAT layer.
We have improved GAT to incorporate edge features into the attention mechanism. 
This improvement can be seen in Line 5 of the algorithm. 
Unlike the standard GAT method that does not utilize the features of edges for computing the attention coefficient, the computation of attention coefficients for each edge of a node in NEGAT involves concatenating the previous layer embeddings of the source node, target node, and the edges between them. This concrete concatenated function is shown as follows:
\begin{equation}
    w^{k}_{vu}=a(\boldsymbol{W^{k}}\vec{h}_{v}^{k-1} || \boldsymbol{W^{k}}\vec{e}_{vu}^{k-1} || \boldsymbol{W^{k}}\vec{h}_{u}^{k-1})
\end{equation}
, where $k$ represents the current layer, and $k-1$ denotes the previous layer, and $\Vert$ represents the feature concatenation operation.
Attention mechanism $a$ is a single-layer feed-forward neural network, where $\boldsymbol{W^{k}}$ represents the weight matrix of current layer $k$ for the input linear transformation. 
$\vec{h}_{v}^{k-1}$ and $\vec{h}_{u}^{k-1}$ denote the previous layer embedding of the central node $v$ and of node $u$, one of the neighboring nodes of $v$, respectively. 
$\vec{e}_{vu}^{k-1}$ refers to the previous layer embedding of edge between the node $v$ of the node $u$.
In Line 7, the normalized attention coefficients $\alpha$ of the node $v$ are calculated similarly to GAT, using the softmax function on $w^{k}_{vu}$ of all adjacent edges of $v$. 
Once $\alpha$ is obtained, Line 8 executes the aggregation process for the node $v$.
The aggregation formula is given as follows: 
\begin{equation}
    \vec {h}_{v}^{k}\longleftarrow\sigma(\sum_{u \in \mathcal{N}_{v}}\alpha^{k}_{vu}\boldsymbol{w}_{\mathbf{1}}^{k}(\vec{h}_{v}^{k-1}||\vec{e}_{vu}^{k-1}||\vec{h}_{u}^{k-1}))
\label{NEGATAggregation}
\end{equation}
, where $h_{v}^{k}$ denotes the aggregated messages, incorporating the non-linearity ${\boldsymbol{\sigma}}$ and the weight matrix $\boldsymbol{w}_{\mathbf{1}}^{k}$. 
The normalized attention coefficients $\alpha_{vu}$ are utilized to calculate a linear combination of the embeddings of the node $v$ and its neighbor nodes $u$ along with their corresponding edges.

The final embedding of graph $G$ will be used as input of the NEGSC model, where the embeddings of nodes are essentially their embeddings at the final layer $K$ (at Lines 11-13), and the embeddings of edges are the concatenation of that of their two end nodes (at Lines 14-16). 

\begin{figure*}[h]
    \includegraphics[width=\textwidth]{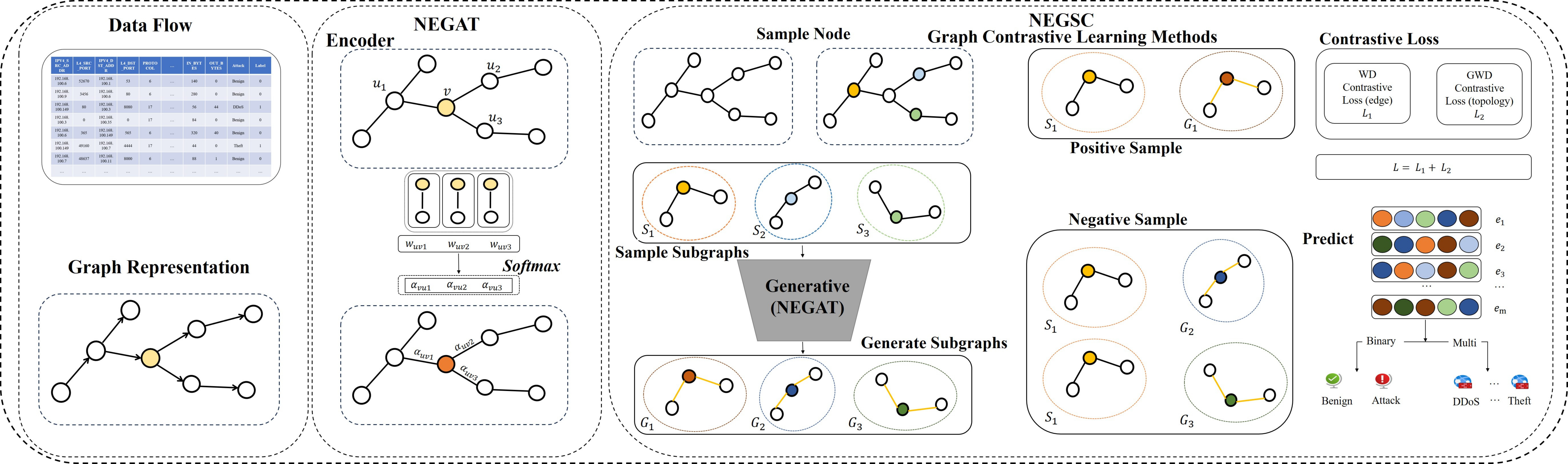}
    \caption{The architecture of our method. We first employ the encoder to obtain the graph embedding. Then, we obtain the subgraphs for every sampled center nodes. Next, we use the generation module to generate the contrastive sample of each sampled subgraph, and pair the sampled and generated subgraphs with the same center node as the positive samples while the subgraphs with the different center nodes as the negative samples. Finally, we use the proposed loss function to measure the distance between the topological structures of subgraph pairs in the positive and negative samples.}
    \label{NEGSC}
\end{figure*}

\subsection{ The Proposed NEGSC Model}
Due to the increasing frequency of attack updates, numerous new or unknown attacks have been arising and need to be responded to in time. 
To augment the adaptability of NIDS to the network environment, we introduce a self-supervised GNN framework NEGSC based on a graph contrastive learning method GSC, for distinguishing normal network traffic flows and different types of malicious network flows. 
In this subsection, we first give a brief description of the original GSC, and then present NEGSC, especially its improvements on GSC, that enables our model for multiclass classification task in NetFlow-based NIDSs databases under unsupervised manner.

\subsubsection{ GSC }
Generative Subgraph Comparison (GSC) is a self-supervised method for graph representation learning that captures the local structure of graph by generating contrastive samples \cite{Y. Han}.  
The method utilizes a contrastive learning framework for adaptive subgraph generation to capture the intrinsic graph structure, and employs optimal transmission distance as a similarity measure between subgraphs.
Given a graph and its embedding, GSC first samples center nodes, and constructs subgraphs around center nodes by the breadth first search (BFS). 
Notice that the path between two nodes in subgraph may cross multiple hops.
Then a generation module is proposed. 
For each sampled subgraph, the module adaptively interpolates new nodes by learning the relationship weights between nodes in subgraph, and automatically generate the edges between the interpolated nodes according to the similarities between nodes.
It next pairs the sampled and generated subgraph with the same center node as the positive samples, and that with different center node as the negative samples.
Finally, two optimal transmission distances, Wasserstein distance and Gromov-Wasserstein distance, are used to construct a structured comparison loss. 
GSC is a general learning method, that mainly focuses on node representation in the graph. 
It lacks sufficient consideration for application in NIDS, where the core information is on edges and edges represent real network flows rather than relationships between hosts.

\subsubsection{ NEGSC }
We now introduce NetFlow-Edge Generative Subgraph Contrast (NEGSC), a self-supervised graph representation learning method for identification of malicious attacks and their specific attack types.
As shown in Fig.\ref{NEGSC}, NEGSC follows a similar framework as that of GSC.
We make numerous improvements by fully exploiting the properties of network traffic flows and their topological structure, in order to enable NEGSC for accurately and effectively characterizing the difference between network flows in NIDS. 

After converting NetFlow-based data to graph representation, the proposed NEGAT, that directly works on edge features, is employed by NEGSC as encoder to get the graph embedding, where each node embedding  $\vec{\boldsymbol{z}}_v$ is aggregated from the adjacent edges features by attention mechanism, and each edge embedding $\vec{\boldsymbol{z}}_{vu}$ is concatenated by two end nodes embeddings $\vec{\boldsymbol{z}}_{v} \Vert \vec{\boldsymbol{z}}_{u}$.
NEGAT helps NEGSC in extracting important information from raw features of network flows. 

At the graph sampling step, NEGSC samples center nodes and for each center node, constructs subgraph by it and the same number of its direct neighbor nodes. 
Therefore, in the sample subgraphs, the distance of each path between node and center node is 1, and the longest distance between nodes is no larger than 2. 
Compared with the BFS method used by GSC, our sampling method is simpler and pays more attention to local structure of graph.
It is because that in a network attack scenario, malicious hosts typically launch attacks on adjacent host, and multi-hop attacks only play a small part of all network attacks, especially when the number of hops is larger than 2. 
And in some specific attacks, such as DoS and DDoS attacks, source address forgery is fairly common, and makes it useless or even impossible to trace network flows far from the host.  

At the contrastive subgraph generating step, NEGSC uses a learnable subgraph generation method to adaptively generate contrastive subgraph for each sampled graph, and then pairs the sampled and generated subgraphs with the same center node as the positive samples, and that with different center node as the negative samples.
Firstly, for each node $v$ in any sampled subgraph $S$, NEGSC conducts local structure information interpolation to generate a new node $v'$ for the corresponding generated contrastive subgraph $G$.
Instead of the formula $\boldsymbol{\hat{z}_{v'}}=\sum_{u \in \mathcal{N}_v} \alpha_{vu}\boldsymbol{\vec{z}}_u$ in GSC, we use 
\begin{equation}
\hat {z}_{v'}\longleftarrow\sigma(\sum_{u \in \mathcal{N}_{v}}\alpha_{vu}\boldsymbol{w}(\vec{z}_{v}||\vec{z}_{vu}||\vec{z}_{u}))
\end{equation}
in NEGSC as interpolation formula, that is similar to Equation (\ref{NEGATAggregation}), and can utilize the embeddings of both edges and nodes more accurately. 
Here, $\hat{z}_{v'}$ is the interpolated embedding of node $v'$ in generated subgraph $G$, $\alpha_{vu}$ is the learned attention coefficient between nodes $v$ and $u$, and $\vec{\boldsymbol{z}}_v$, $\vec{\boldsymbol{z}}_u$ and $\vec{\boldsymbol{z}}_{vu}$ are the input embeddings of nodes $u$, $v$ and edge $e_{vu}$, respectively.
Secondly, GSC directly generates edges between new nodes in generated graph if and only if there exist edges between the corresponding nodes in sampled graph, i.e., $e_{v'u'}$ exists in $G$ IFF $e_{vu}$ exists in $S$.
The embedding of each edge $\boldsymbol{\hat{z}}_{v'u'}$ is the concatenation of its two end nodes representations  $\boldsymbol{\hat {z}}_{v'}||\boldsymbol{\hat{z}}_{u'}$.
NEGSC avoids the time-consuming method in GSC, that generates edges according to the cosine similarity between new nodes. 
The reason is that in NIDS, edge represents network traffic flow between nodes, and the representation similarity between two nodes does not imply they communicate with each other. 

Then, we come to the loss function in NEGSC.
A new loss function based on the optimal transport distance is proposed in order to accurately characterize the geometric difference between the sampled subgraph and the generated subgraph in the positive and negative samples, in terms of the edge and graph topological structure.
Specifically, we utilize the Wasserstein Distance to measure the contrastive loss on edge features, and utilize the Gromov-Wasserstein Distance to measure the contrastive loss on the topological structure of graph.

Wasserstein Distance (WD) \cite{G. Peyré} is a common method used for matching two discrete distributions, e.g., two sets of node (or edge) embeddings.
Here, we use WD to evaluate the similarity between the edges in the sampled and generated subgraph of each positive or negative sample.
Let $\mu = \{ \mu_1, \cdots, \mu_m \}$ and $\nu = \{ \nu_1, \cdots, \nu_m \}$ be discrete distributions of two subgraphs $S$ and $G$, respectively, where $\sum_{i=1}^{m} \mu_i = 1$ and $\sum_{j=1}^{m} \nu_j = 1$, and $m$ is the number of edges in subgraph (Recall that $S$ and $G$ have the same number of edges and nodes due to the graph sampling step and the contrastive subgraph generating step in NEGSC). The WD between two distributions $\mu$ and $\nu$ is defined as:

\begin{equation}
    WD{(S, G)}=\min_{\boldsymbol{T}\in\pi({\mu},{\nu})} \sum_{i=1}^{m} \sum_{j=1}^{m}\boldsymbol{T}_{ij} c({\vec{z}}_{e_{i}},{\hat{z}}_{e_{j}}).
\end{equation}
$\pi({\mu}, {\nu})$ denotes all the joint distributions of the edges in subgraphs $G$ and $S$. 
$c({\vec{z}}_{e_{i}},{\hat{z}}_{e_{j}})$ is the cost function evaluating the distance between edge $e_i$ in subgraph $S$ and edge $e_j$ in subgraph $G$, where ${\vec{z}}_{e_{i}}$ and ${\hat{z}}_{e_{j}}$ represent the embeddings of $e_i$ and $e_j$, respectively.
The matrix $\boldsymbol{T}$ is the transport plan, and $\boldsymbol{T}_{ij}$ denotes the amount of mass shifted from $e_i$ and $e_j$. 
Using the same cost function in GSC for $c({\vec{z}}_{e_{i}},{\hat{z}}_{e_{j}})$, the WD between $\mu$ and $\nu$ can be efficiently achieved by the methods proposed in \cite{chen2020OT}.

Gromov-Wasserstein Distance (GWD) \cite{G. Peyré and M. Cuturi} is commonly used to evaluate the distance between pairs of nodes within graph, and further to measure the differences in these distances across the subgraphs.  
Here, we use GWD to measure the similarity between topological structures of two subgraphs in each positive or negative sample.
Similarly, $\mu = \{ \mu_1, \cdots, \mu_n \}$ and $\nu = \{ \nu_1, \cdots, \nu_n \}$ denote distributions of two subgraphs $S$ and $G$, respectively, where $\sum_{i=1}^{n} \mu_i = 1$ and $\sum_{j=1}^{n} \nu_j = 1$, and $n$ is the number of nodes in subgraph.
The GWD of two discrete distributions $\mu$ and $\nu$ is defined as:
\begin{equation}
    GWD{(S,G)}=\min_{\boldsymbol{T} \in \pi({\mu},{\nu})} \sum_{v, u,v',u'}\boldsymbol{T}_{vv'}\boldsymbol{T}_{uu'}
    \hat{c}(\vec{z}_{v},\hat{z}_{v'},\vec{z}_{u},\hat{z}_{u'}).
\end{equation}
Edges $e_{vu}$ and $e_{v'u'}$ respectively exist in the subgraphs $S$ and $G$. 
$\hat{c}(\vec{z}_{v},\hat{z}_{v'},\vec{z}_{u},\hat{z}_{u'}) = \|c(\boldsymbol{z}_{v},\boldsymbol{z}_{u})-c(\boldsymbol{z}_{v'},\boldsymbol{z}_{u'})\|_2$ is the cost function measuring the topological structure distance between two pairs of nodes in different subgraphs, where we used the same function $c( \cdot , \cdot )$ in GSC to denote the distance between the two nodes within a subgraph.

WD-Based Contrastive Loss is:
\begin{multline}
    \mathcal{L}_{edges}=\frac{-1}{N(M+1)}\sum_{i=1}^N[\log(\exp(-WD(S_i,G_i)/\tau))\\
    +\sum_{ni \in \{1, ..., M\} - \{ i \} } \log(1 - \exp(-WD(S_i,G_{ni})/\tau))]
\end{multline}
for measuring the similarity of edges,
and GWD-Based Contrastive Loss is:
\begin{multline}
    \mathcal{L}_{topology}=\frac{-1}{N(M+1)}\sum_{i=1}^N[\log(\exp(-GWD(S_i,G_i)/\tau))\\
    +\sum_{ni \in \{1, ..., M\} - \{ i \} } \log(1-\exp(-GWD(S_i,G_{ni})/\tau))]
\end{multline}
for measuring the similarity of graph topological structure.
$N$ is the number of sampled subgraphs, $M$ is the number of negative samples of each subgraph, and $\tau$ is a temperature parameter. 
$(S_i,G_i)$ denotes the positive sample subgraphs pair, and $(S_i,G_{ni})$ denotes the negative sample subgraphs pair. 
Essentially, the GWD-based contrastive loss in NEGSC is the same as that in GSC, although GSC calls it edge loss while NEGSC calls it topological loss.

Finally, the loss function of NEGSC is defined as follow:
\begin{equation}
    \mathcal{L}= \mathcal{L}_{edges} + \mathcal{L}_{topology}.
\end{equation}
Since the critical information is on network flows in netflow-based NIDS, compared with the function in GSC, our function focuses on the edge embeddings, i.e., the properties of network flows, and the relationships among the adjacent edges, while ignoring the node embeddings.

\subsection{ The Time Complexity Analysis}
The running time of our model is determined by its two modules, NEGAT and NEGSC. 
We first consider the time complexity of NEGAT.
In a single attention head, each node need to consider its adjacent edges along with their nodes for computing attention coefficients, and hence each edge is counted twice.
Therefore, the attention coefficient computing step takes time $O(2m)$, where $m$ is the number of edge in the graph.
Similarly, the aggregating step also takes time $O(2m)$.
Let $I$ and $K$ be the number of attention heads and of the layers, respectively, the time complexity of NEGAT is $O(4KIm)$.
In the NEGSC module, the first sampling step takes time $O(n)$, where $n$ is the number of nodes in the graph, because the sampled subgraphs have no common nodes.
The generation module interpolates new node for each node in sampled subgraphs, which can be regarded as an aggregating step in NEGAT and thus takes time $O(2Im)$, and constructs edges between node, which takes time $O(n)$.
The final step gets the value of the loss function by a time-consuming operation of graph matching.
However in our setting of method, the sampled subgraph and the generated subgraph consist of a fixed number of nodes and edges, and the number $N$ and $M$ of positive and negative samples are also fixed.
Hence the matching operation between subgraphs can be done in constant time. 
Therefore in all, the time complexity of the proposed model is $O(4KIm + 2Im + 2n)$, which is an efficient linear time on the number of nodes and edges in the graph.

\subsection{ Training }
Each dataset is divided into a training set and a testing set in the same proportion. 
Before the training of our self-supervised model NEGSC, all data are converted into graph representation, and are fed into the encoder NEGAT to get a powerful embedding. 
Note that we use Layer $K$ = 1 in NEGAT, due to consideration of the complexity and computational efficiency of our model. 
Detailed information on hyperparameters is provided in Table \ref{tab:Values}.  
In the loss function, we use the binary cross-entropy (BCE) function and the cross-entropy (CE) function. 
These are commonly used loss functions that can effectively measure the gap between the predicted and true values of the model. 
In order to optimize the model parameters, back-propagation gradient descent is used. 
In addition, we use the Adam optimizer and set the learning rate in the training set to 0.001.

\begin{table}[h!]
    \footnotesize
    \centering
    \captionsetup{
        width=0.267\textwidth, 
    singlelinecheck=false, 
    justification=raggedright} 
    \caption{Hyperparameter.}
    \renewcommand{\familydefault}{\rmdefault}
    \label{tab:Values}
    \begin{tabular}{lcll}
        \toprule[0.1em]
        Hyperparameter & & & Value \\
        \midrule[0.1em]
        Layer $K$  & & & 1 \\
        Attention heads $I$ & & & 3 \\
        Learning Rate & & & [2e-2,1e-3] \\
        Activation Func & & & RELU \\
        Loss Func& & & BCE/CE \\
        Optimizer& & & Adam \\
        \bottomrule[0.1em]
    \end{tabular}
\end{table}

\begin{table*}[h!]
    \footnotesize
    \centering
    \captionsetup{
        singlelinecheck=false, 
        justification=raggedright,
        width=0.655\textwidth
    }
    \caption{Statistics of datasets used in our experiments.}
    \label{sum datasets}
    \begin{tabular}{|c|c|c|c|c|}
    \hline
    \multirow{2}{*}{\textbf{Dataset}} & \multirow{2}{*}{\textbf{Attribute}} & \multirow{2}{*}{\textbf{Classe}} & \multicolumn{2}{c|}{\textbf{Sample}} \\ 
    \cline{4-5}
    & & & version1 & version2 \\ 
    \hline
    \multirow{7}{*}{\shortstack{ NF-Bot-IoT \cite{dataset1} \\ NF-BoT-IoT-v2 \cite{dataset2} }} & \multirow{2}{*}{Label} & Normal & 13859 & 4051 \\ 
    & & Attack& 586241 & 1128853 \\ 
    \cline{2-5}
    & \multirow{5}{*}{Attack} & Benign & 13859 & 4051 \\ 
    & & Theft & 1909 & 73 \\ 
    & & DoS & 56833 & 500195 \\ 
    & & DDoS & 56844 & 549955 \\ 
    & & Reconnaissance & 470655 & 78630 \\ 
    \hline
    \multirow{17}{*}{\shortstack{ NF-CSE-CIC-IDS2018 \cite{dataset1} \\ NF-CSE-CIC-IDS2018-v2 \cite{dataset2} }} & \multirow{2}{*}{Label} & Normal & 737320 & 831778 \\
    & & Attack & 101920 & 112908 \\ 
    \cline{2-5}
    & \multirow{15}{*}{Attack} & Benign & 737320 & 831778 \\
    & & DDOS attack-HOIC & 23 &  54043 \\
    & & DoS attacks-Hulk & 10814 & 21632 \\
    & & DDoS attacks-LOIC-HTTP & 37820 & 15365 \\
    & & Bot & 1568 & 7155 \\
    & & Infilteration & 6207 & 5818\\
    & & SSH-Bruteforce & 5654 & 4749 \\
    & & DoS attacks-GoldenEye & 3285 & 1386 \\
    & & FTP-BruteForce & 11602 & 1297 \\
    & & DoS attacks-SlowHTTPTest & 10555 & 706 \\
    & & DoS attacks-Slowloris & 2282 & 476 \\
    & & Brute Force -Web & 261 & 107 \\
    & & DDOS attack-LOIC-UDP & 167 & 106 \\
    & & Brute Force -XSS & 174 & 46 \\
    & & SQL Injection & 4 & 22 \\
    \hline
    \end{tabular}
\end{table*}

\section{Experiments and Results}
\label{4}
In this section, we conduct extensive experiments to evaluate performance of our method under a self-supervised manner for both binary and multiclass classification tasks.
Besides, ablation experiments are provided to seprately illustrate the effectiveness of our propsoed encoder NEGAT and self-supervised framework NEGSC.

\subsection{ Setup }
Our model is implemented using Python, Pytorch \cite{A. Paszke} and DGL \cite{M. Wang}.
The experiments are conducted on a single NVIDIA GeForce RTX 4090 with 24GB of GPU memory.
We use the hold-out method \cite{C. Bishop} to divide the datasets.
First randomly choose data from datasets in a fixed proportion.
Then in the training phase, 70\% of the data are sampled for training. 
This data is preprocessed and fed into the model, which is iteratively optimized so that the model can learn from it and extract useful information and knowledge. The remaining 30\% of the data is used for testing and performance evaluation.
Code to reproduce our experiments is available at \url{https://github.com/renj-xu/NEGSC}.

\subsection{ Datasets }
In the study, we select four public NetFlow-based  benchmark NIDS datasets: NF-BoT-IoT \cite{dataset1}, NF-BoT-IoT-v2 \cite{dataset2}, NF-CSE-CIC-IDS2018 \cite{dataset1} and NF-CSE-CIC-IDS2018-v2 \cite{dataset2}.
The NF-BoT-IoT and the NF-BoT-IoT-v2 are datasets on network traffic of IoT.
The NF-CSE-CIC-IDS2018 and the NF-CSE-CIC-IDS2018-v2 are network intrusion detection datasets collected on server and host.
Table \ref{sum datasets} specifies the details of the four datasets, including attack categories, predictable attributes, as well as the amount of data we used.
All these datasets support binary and multiclass classification scenarios. 

The NF-BoT-IoT was generated from the original dataset BoT-IoT \cite{N. Koroniotis} and made up of 8 basic NetFlow-based features. 
It contains 600,100 network flows (entries) in total, out of which 13,859 are benign samples, and 586,241 are attack samples, uneven distributed among 4 attack types.
The NF-BoT-IoT-v2 was generated by the NF-BoT-IoT, that expands the number of network flow features from 8 to 39, and increases the number of network flows. 
The number of network flows after increasing is 37,763,497, of which 135,037 are benign samples, and 37,628,460 are attack samples. 

The NF-CSE-CIC-IDS2018 was generated from the original dataset CSE-CIC-IDS2018\cite{I. Sharafaldin} and made up of 8 basic NetFlow-based features. 
It contains 8,392,401 total network flows, out of which 7,373,198 are benign, and 1,019,203 are attack samples,
uneven distributed among 14 attack types.
The NF-CSE-CIC-IDS2018-v2 was generated by the NF-CSE-CIC-IDS2018, that also expands the number of network flow features from 8 to 39, and increases the number of network flows. 
The number of network flows after increasing is 18,893,708, out of which 16,635,567 are benign, and 2,258,141 are attack samples. 

The NF-CSE-CIC-IDS2018 and the NF-CSE-CIC-IDS2018-v2 datasets show a significantly smaller number of attack samples than that of benign samples.
This situation reflects a common phenomenon in the real world, where anomalous behaviors (e.g., attacks) are usually less than benign behaviors (e.g., normal network flow) in a network environment. 

Considering that the NF-BoT-IoT-v2, NF-CSE-CIC-IDS2018 and NF-CSE-CIC-IDS2018-v2 datasets are too large to process completely in our experiment environment, we randomly selected 3\%, 10\%, and 5\% of the original data, respectively.
Each dataset has two attributes, providing information and diversity of network flows. 
The attribute ``Label" helps us evaluate the performance of the model in binary classification tasks, where 0 represents benign behavior and 1 represents anomalous behavior;
the other attribute ``Attack" helps us evaluate our model performance in multiclass classification tasks, which is assigned difference values representing a specific type of attack.
See table \ref{sum datasets} for more details.
   
\begin{table*}[h!]
    \footnotesize
    \begin{minipage}[t]{.5\textwidth}  
        \centering
        \captionsetup{
            singlelinecheck=false, 
            justification=raggedright,
            width=0.816\textwidth
        }
        \caption{On dataset NF-CSE-CIC-IDS2018 binary classification results.} 
        \label{model-b-NF-CSE-CIC-IDS2018}
        \begin{tabular}{ccccc}
            \toprule[0.1em]
            \textbf{Model} & \textbf{Acc} & \textbf{Precision} & \textbf{Recall} & \textbf{F1} \\
            \midrule[0.1em]
            \multirow{2}{*}\textbf{\shortstack { Anomal-E \\ (0\%)-average}} & 82.65\% & 89.12\% & 82.65\% & 84.73\% \\
            &&&&\\
            \multirow{2}{*}\textbf{\shortstack { Anomal-E \\ (0\%)-max}} & 87.32\% & 89.35\% & 87.32\% & 88.12\% \\
            &&&&\\
            \multirow{2}{*}\textbf{\shortstack { Anomal-E \\ (4\%)-average}} & 79.60\% & 81.95\% & 79.60\% & 80.54\% \\
            &&&&\\
            \multirow{2}{*}\textbf{\shortstack { Anomal-E \\ (4\%)-max}} & 85.46\% & 88.65\% & 85.46\% & 86.68\% \\
            &&&&\\
            \textbf{Ours}  & \textbf{97.66\%} & \textbf{96.75\% }& \textbf{96.77\%} & \textbf{96.76\%} \\
            \bottomrule[0.1em]
        \end{tabular}
    \end{minipage}%
    \begin{minipage}[t]{.5\textwidth}  
        \centering
        \captionsetup{
            singlelinecheck=false, 
            justification=raggedright,
            width=0.816\textwidth
        }
        \caption{On dataset NF-BoT-IoT binary classification results.} 
        \label{model-b-NF-BoT-IoT}
        \begin{tabular}{ccccc}
            \toprule[0.1em]
            \textbf{Model} & \textbf{Acc} & \textbf{Precision} & \textbf{Recall} & \textbf{F1} \\
            \midrule[0.1em]
            \multirow{2}{*}\textbf{\shortstack { Anomal-E \\ (0\%)-average}} & 20.67\% & 63.80\% & 20.67\% & 28.05\% \\
            &&&&\\
            \multirow{2}{*}\textbf{\shortstack { Anomal-E \\ (0\%)-max}} & 42.34\% & 96.58\% & 42.34\% & 57.22\% \\
            &&&&\\
            \multirow{2}{*}\textbf{\shortstack { Anomal-E \\ (4\%)-average}}  & 22.10\% & 91.88\% & 22.10\% & 34.40\% \\
            &&&&\\
            \multirow{2}{*}\textbf{\shortstack { Anomal-E \\ (4\%)-max}} & 23.28\% & 92.20\% & 23.28\% & 35.97\% \\
            &&&&\\
            \textbf{Ours} & \textbf{98.77\%} & \textbf{98.70\%} & \textbf{98.77\%} & \textbf{98.59\%} \\
            \bottomrule[0.1em]
        \end{tabular}
    \end{minipage}

    \vspace{0.5cm}  

    \begin{minipage}[t]{.5\textwidth}  
        \centering
        \captionsetup{
            singlelinecheck=false, 
            justification=raggedright,
            width=0.816\textwidth
        }
        \caption{On dataset NF-CSE-CIC-IDS2018-v2 binary classification results.} 
        \label{model-b-NF-CSE-CIC-IDS2018-v2}
        \begin{tabular}{ccccc}
            \toprule[0.1em]
            \textbf{Model} & \textbf{Acc} & \textbf{Precision} & \textbf{Recall} & \textbf{F1} \\
            \midrule[0.1em]
            \multirow{2}{*}\textbf{\shortstack { Anomal-E \\ (0\%)-average}} & 97.83\% & 97.86\% & 9783\% & 97.75\%\\
            &&&&\\
            \multirow{2}{*}\textbf{\shortstack { Anomal-E \\ (0\%)-max}} & \textbf{97.87\%} & \textbf{97.90\%} & \textbf{97.87\%} & \textbf{97.79\%} \\
            &&&&\\
            \multirow{2}{*}\textbf{\shortstack { Anomal-E \\ (4\%)-average}} & 96.14\% & 95.66\% & 94.45\% & 94.74\% \\
            &&&&\\
            \multirow{2}{*}\textbf{\shortstack { Anomal-E \\ (4\%)-max}}  & 97.02\% & 96.98\% & 97.02\% & 96.90\% \\
            &&&&\\
            \textbf{Ours} & 97.76\% & 97.78\% & 97.76\% & 97.67\% \\
            \bottomrule[0.1em]
        \end{tabular}
    \end{minipage}%
    \begin{minipage}[t]{.5\textwidth}  
        \centering
        \captionsetup{
            singlelinecheck=false, 
            justification=raggedright,
            width=0.816\textwidth
        }
        \caption{On dataset NF-BoT-IoT-v2 binary classification results.}  
        \label{model-b-NF-BoT-IoT-v2}
        \begin{tabular}{ccccc}
            \toprule[0.1em]
            \textbf{Model} & \textbf{Acc} & \textbf{Precision} & \textbf{Recall} & \textbf{F1} \\
            \midrule[0.1em]
            \multirow{2}{*}\textbf{\shortstack { Anomal-E \\ (0\%)-average}} & 38.75\% & 99.42\% & 38.75\% & 55.37\% \\
            &&&&\\
            \multirow{2}{*}\textbf{\shortstack { Anomal-E \\ (0\%)-max}} & 43.51\% & \textbf{99.44\%} & 43.51\% & 60.28\% \\
            &&&&\\
            \multirow{2}{*}\textbf{\shortstack { Anomal-E \\ (4\%)-average}} & 29.68\% & 99.26\% & 29.68\% & 44.82\% \\
            &&&&\\
            \multirow{2}{*}\textbf{\shortstack { Anomal-E \\ (4\%)-max}} & 39.82\% & 99.21\% & 39.82\% & 56.64\% \\
            &&&&\\
            \textbf{Ours}& \textbf{99.64\%} & 99.29\% & \textbf{99.64\%}& \textbf{99.46\%} \\
            \bottomrule[0.1em]
        \end{tabular}
    \end{minipage}
\end{table*}

\subsection{ Evaluation Metrics }
To evaluate the performance of our proposed NIDS model, we select four metrics, namely, Accuracy (Acc) rate, Recall rate, Precision rate, and F1 Score. 
These metrics have been widely utilized in numerous studies \cite{M. Sarhan,D.M.W. Powers}, that can be used to evaluate the performance of model in both multiclass classification and binary classification.

The calculation of these metrics is based on four counts: true positives ($TP$), the model correctly predicts the positive class; true negatives ($TN$), the model correctly predicts the negative class; false positives ($FP$) The model incorrectly predicts the positive class, and false negatives ($FN$) The model incorrectly predicts the negative class.

Accuracy measures the proportion of correct predictions for all samples.
\begin{equation}
    \mathrm{Acc} = \frac{ TP + TN }{ TP + FP + TN + FN } \times 100\%
\end{equation}

Precision measures the proportion of samples predicted as attacks by the model that are actual attacks.
\begin{equation}
    \mathrm{Precision} = \frac { TP }{ TP + FP } \times 100\%
\end{equation}

Recall reflects the ability of the model to identify attacks and measures the sensitivity of the model to attack behavior.
\begin{equation}
    \mathrm{Recall} = \frac{ TP }{ TP + FN } \times 100\%
\end{equation}

The F1 score is the harmonic mean of Precision and Recall.
\begin{equation}
    \mathrm{F}1 = 2\times\frac{\mathrm{Precision} \times \mathrm{Recall}}{\mathrm{Precision} + \mathrm{Recall}} \times 100\%
\end{equation}

\subsection{ Experimental Results }
The subsection provides extensive experiment results on our proposed method for classifying network flows in NIDS under self-supervised manner.
Binary classification experiments are first provided to illustrate the ability of distinguishing between benign and malicious network flows. 
Then, more complex multiclass classification experiments are conducted to evaluate the ability of identifying different attack types. 
Additionally, ablation experiments are given to evaluate the effectiveness and contribution of each component (NEGAT and NEGSC) in our method.
We report that the results by excerpting from the corresponding papers or by running the code when available.

\subsubsection{ Binary Classification Results }
In the binary classification experiments, we compare the proposed method with Anomal-E \cite{E. Caville} and TS-IDS \cite{H. Nguyen}. 
Anomal-E is the only GNN-based self-supervised approach for binary classification in NIDS. 
TS-IDS is a latest GNN-based supervised approach, that incorporates a self-supervised module.

The Anomal-E method uses four traditional anomaly detection algorithms to get the results: principal component analysis-based anomaly detection, isolation forest-based anomaly detection, clustering-based local outlier factor and histogram-based outlier score.
For the sake of simplicity, our method is compared with the max and average value of the four anomaly detection algorithms results in Anomal-E.
Anomal-E (0\%) indicates that all sampled network flows in the training sets are benign,
and Anomal-E (4\%) indicates that 4\% of sampled network flows are attacks.
It is worth noting that in our method, the sampled data in the training and testing sets follow the same distribution in the corresponding dataset. 

Table \ref{model-b-NF-CSE-CIC-IDS2018} and Table \ref{model-b-NF-BoT-IoT} show the binary classification results of our method and Anomal-E on the NF-CSE-IDS2018 and NF-BoT-IoT datasets, respectively, in terms of Acc, Precision, Recall and F1. 
Compared with the Anomal-E, our method achieves the best values in all four metrics, and significant outperforms on the NF-BoT-IoT dataset.
Table \ref{model-b-NF-CSE-CIC-IDS2018-v2} and Table \ref{model-b-NF-BoT-IoT-v2} report the comparison results on the NF-CSE-IDS2018-v2 and NF-Bot-IoT-v2 datasets, respectively.
On the NF-CSE-IDS2018-v2 dataset, our method achieves Acc of 97.76\%, Precision of 97.78\%, Recall of 97.76\% and F1 of 97.67\%, which are very close to those of Anomal-E.
On the NF-BoT-IoT-v2 dataset, our method outperforms Anomal-E in all the metrics except Precision, whose value is also very close.
It can be found that our method achieves the better predictive performance than Anomal-E, especially on the datasets in which the number of benign flows is much smaller than that of attack flows.
It is probably because Anomal-E needs sufficient benign flows for training, while our method, based on Generative Subgraph Comparison, is more adapted to the situation.

To further validate the performance of our model, a comparison with the supervised GNN-based method TS-IDS is conducted.
As shown in Table \ref{MODEL WITH TS-IDS}, on the NF-BoT-IoT, NF-BoT-IoT-v2, and NF-CSE-CIC-IDS2018 datasets, our method is better than TS-IDS in terms of the four metrics, and even has the gain of 9\% improvement in Acc and Recall on NF-BoT-IoT-v2.
On the NF-CSE-CIC-IDS2018-v2 dataset, TS-IDS slightly exceeds our method of less than 2\% in all metrics.

Consequently, the results show that our method has an excellent capability of binary classification compared with the latest methods, which can effectively distinguish normal and malicious network flows without labeled samples. 

\begin{table}[h!]
    \footnotesize
    \centering
    \captionsetup{
    singlelinecheck=false, 
    justification=raggedright,
    width=0.484\textwidth
    }
    \caption{Comparison of our method and TS-IDS in binary classification results.}
    \label{MODEL WITH TS-IDS}
        \begin{tabular}{cccccc}
            \toprule[0.1em]
            \multirow{2}{*}{\textbf{Model}} & \multirow{2}{*}{\textbf{Datasets}} & \multicolumn{4}{c}{\textbf{Metric}} \\
            \cline{3-6}
             & & \bf{Acc} & \bf{Precision} & \bf{Recall} & \bf{F1} \\
            \midrule[0.1em]
            TS-IDS & \multirow{2}{*}{NF-BoT-IoT} & 92.67\% & 98.16\% & 92.67\% & 94.67\% \\
            Ours & & \bf{98.77\%} & \bf{98.70\% }& \bf{98.77\%} & \bf{98.59\% } \\
            TS-IDS & \multirow{2}{*}{\makecell{NF-BoT-IoT\\-v2}} & 90.50\% & 98.12\% & 90.50\% & 93.45\% \\
            Ours &  & \bf{99.64\%} & \bf{99.29\% }& \bf{99.64\%} & \bf{99.46\% } \\
            TS-IDS & \multirow{2}{*}{\makecell{NF-CSE-CIC-\\IDS2018}} & 89.58\% & 93.36\% & 89.58\% & 90.66\% \\
            Ours &  & \bf{96.77\%} & \bf{96.75\%} & \bf{96.77\%} & \bf{96.76\% } \\
            TS-IDS & \multirow{2}{*}{\makecell{NF-CSE-CIC-\\IDS2018-v2}} & \bf{99.95\%} & \bf{99.02\%} & \bf{98.85\% }& \bf{99.95\% }\\
            Ours &  & 97.76\% & 97.78\% & 97.76\% & 97.67\%  \\
            \bottomrule[0.1em]
        \end{tabular}
    \end{table}

\subsubsection{ Multiclass Classification Results }

Multiclass classification experiments are conducted to demonstrate the ability of the proposed method for distinguishing different types of network flows without labels. 

Table \ref{MODEL multi IDS2018} presents the detailed values of Recall and F1 achieved on two datasets, NF-CSE-CIC-IDS2018 and NF-CSE-CIC-IDS2018-v2. 
Recall reaches 95.79\% and 97.85\% for two datasets respectively.
Especially for the Dos attacks-Hulk, SSH-Bruteforcce, and FTP-Bruteforce in the NF-CSE-CIC-IDS2018-v2 dataset, the high Recall values of 99.60\%, 99.93\% and 99.49\%, respectively, are achieved. 
The detection rates for DoS attacks-SlowHTTPTest, Brute Force-XSS and SQL Injection are 0 on both datasets. 
It may be caused by lack of sufficient corresponding flow samples due to the data unbalance in datasets (refer to Table \ref{sum datasets}).
Table \ref{MODEL multi NF-BoT-IoT} reports the result of Recall and F1 on the NF-BoT-IoT and NF-BoT-IoT-v2 datasets, respectively, where Recall achieves 83.06\% and 95.16\%, and F1 reaches 82.70\% and 95.15\%. 
The difference between our method performances on two datasets may rely on that NF-BoT-IoT-v2 provides more attribute features than NF-BoT-IoT.

\begin{table}[h!]
    \footnotesize
    \centering
    \caption{Multiclass classification results of our method on NF-CSE-CIC-IDS2018 and NF-CSE-CIC-IDS2018-v2.} 
    \label{MODEL multi IDS2018}
    \begin{tabular}{lcccc}
        \toprule[0.1em]
        \multicolumn{1}{c}{\textbf{Class Name}}
        & \multicolumn{2}{c}{\textbf{\makecell{NF-CSE-CIC-\\IDS2018}}} 
        & \multicolumn{2}{c}{\textbf{\makecell{NF-CSE-CIC-\\IDS2018-v2}}}  \\
        \midrule[0.1em]
        & \textbf{Recall} & \textbf{F1} & \textbf{Recall} & \textbf{F1} \\
        
        Benign & 98.95\% & 98.36\% & 99.72\% & 98.89\% \\
        
        DDOS attack-HOIC & 00.00\% & 00.00\% & 90.17\% & 94.00\% \\
        
        DoS attacks-Hulk & 87.11\% & 86.65\% & 99.60\% & 98.72\% \\
        
        DDoS attacksLOICHTTP & 99.83\% & 99.77\% & 87.00\% & 91.25\% \\
        
        Bot & 00.00\% & 00.00\% & 53.91\% & 64.55\% \\
        
        Infilteration & 00.00\% & 00.00\% & 00.17\% & 00.33\% \\
        
        SSH-Bruteforce & 88.15\% & 87.53\% & 99.93\% & 98.82\% \\
        
        DoS attacks-GoldenEye & 82.76\% & 64.97\% & 91.35\% & 89.10\% \\
        
        FTP-BruteForce & 82.61\% & 65.66\% & 99.49\% & 78.50\% \\
        
        DoS attacksSlowHTTPTest & 00.00\% & 00.00\% & 00.00\% & 00.00\% \\
        
        DoS attacks-Slowloris & 01.17\% & 02.27\% & 00.00\% & 00.00\% \\
        
        Brute Force -Web & 01.28\% & 02.38\% & 00.00\% & 00.00\% \\
        
        DDOS attack-LOIC-UDP & 74.00\% & 69.81\% & 65.62\% & 79.25\% \\
        
        Brute Force -XSS & 00.00\% & 00.00\% & 00.00\% & 00.00\% \\
        
        SQL Injection & 00.00\% & 00.00\% & 00.00\% & 00.00\% \\
        
        \textbf{Weighted Average} & \bf{95.79\%} & \bf{94.79\%} & \bf{97.85\% }& \bf{97.43\%} \\
        \bottomrule[0.1em]
    \end{tabular}
\end{table}

\begin{table}[h!]
    \footnotesize
    \centering
    \captionsetup{
        singlelinecheck=false, 
        justification=raggedright,
        width=0.436\textwidth
    }
    \caption{Multiclass classification results of our method on NF-BoT-IoT and NF-BoT-IoT-v2.} 
    \label{MODEL multi NF-BoT-IoT}
    \begin{tabular}{lcccc}
            \toprule[0.1em]
            \multicolumn{1}{c}{\textbf{Class Name}}
            & \multicolumn{2}{c}{\textbf{NF-BoT-IoT}} 
            & \multicolumn{2}{c}{\textbf{NF-BoT-IoT-v2}} \\
            \midrule[0.1em]
            & \textbf{Recall} & \textbf{F1} & \textbf{Recall} & \textbf{F1} \\

            Benign & 44.76\% & 58.44\% & 61.40\% & 71.83\% \\

            DDoS & 55.92\% & 46.42\% & 97.46\% & 96.58\% \\

            DoS & 26.41\% & 31.35\% & 93.56\% & 94.71\% \\

            Reconnaissance & 94.64\% & 94.33\% & 91.11\% & 89.26 \%\\

            Theft & 00.52\% & 01.04\% & 31.82\% & 24.35\% \\

            \textbf{Weighted Average} & \bf{83.06\%} & \bf{82.70\%} & \bf{95.16\%} & \bf{95.15\%} \\
            \bottomrule[0.1em]
    \end{tabular}
\end{table}

\begin{figure}[h]
    \centering
        \centering
        \includegraphics[height = 0.25 \textheight]{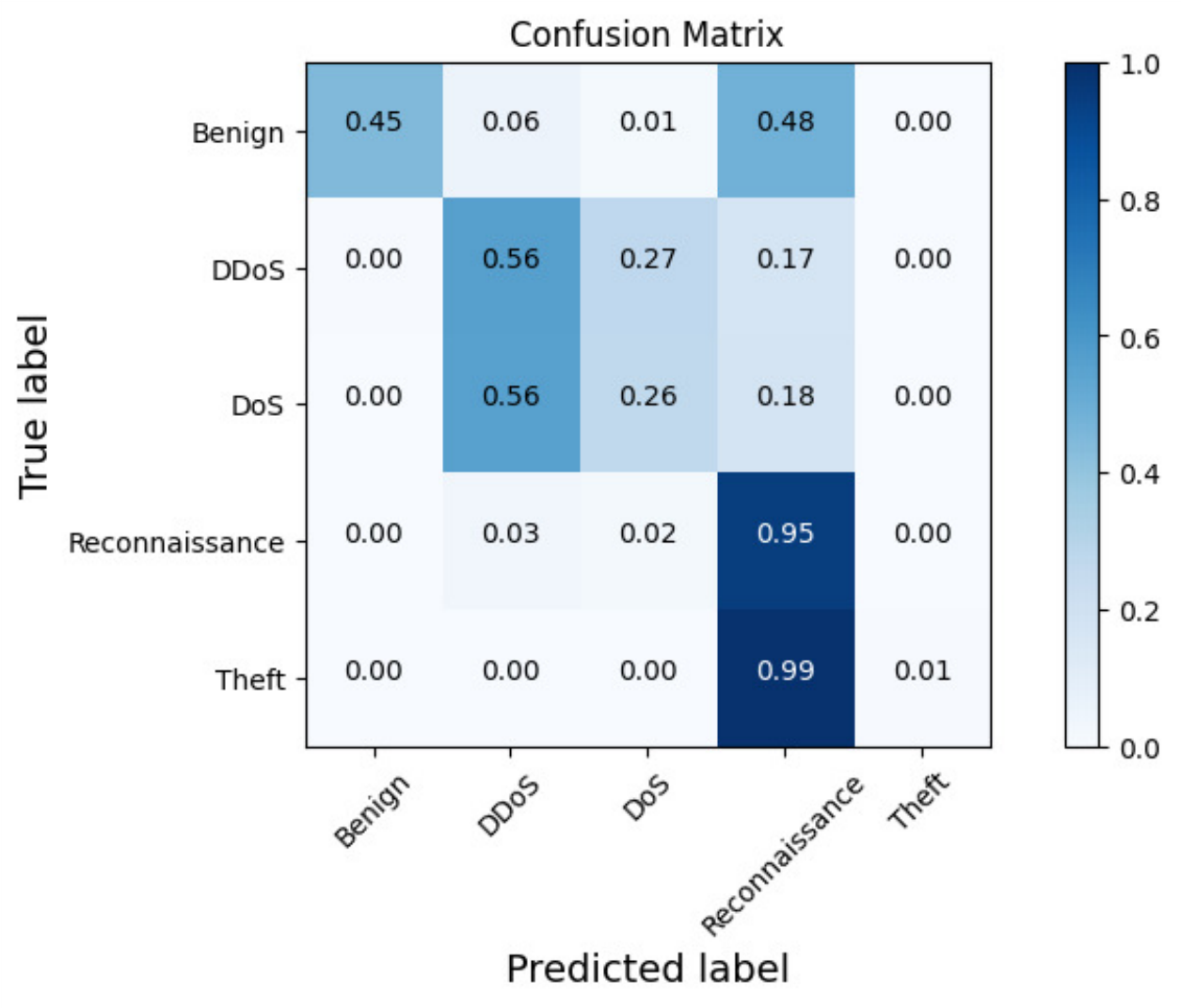}
        \caption{Confusion matrix of multiclass classification results on NF-BoT-IoT.}
        \label{Confusion Matrix}
\end{figure}

As seen from Table \ref{MODEL multi IDS2018} and Table \ref{MODEL multi NF-BoT-IoT}, our proposed method performs well on the NF-CSE-CIC-IDS2018, NF-CSE-CIC-IDS2018-v2 and NF-BoT-IoT-v2 datasets, with Recall and F1 around 95\%.
A relatively poor experimental results are obtained on the NF-BoT-IoT dataset. 
Therefore, we illustrate the confusion matrix for the NF-BoT-IoT dataset in Fig.\ref{Confusion Matrix}.
From the Confusion matrix, it can be found that most theft and benign samples in the NF-BoT-IoT datasets are misclassified as reconnaissance samples. 
This may be related to the similar characteristics of these three types of samples.
Compared to reconnaissance samples that are the most numerous, far more than other categories, the theft and benign samples only occupy a minor fraction of the data in the dataset.
Our method may need more corresponding data for training.
DoS attacks are misclassified as DDoS attacks, probably because DoS samples are most similar to DDoS samples when extracting features.

The experimental result in Table \ref{Proposed Multi results} shows the comparison between the supervised method TS-IDS and our method in Recall and F1 metrics. 
On the NF-BoT-IoT dataset, our method obtains the gain of at least 2\% improvement on both metrics, and achieves Recall of 95.16\% and F1 of 95.15\% on NF-BoT-IoT-v2, closing to that of TS-IDS.
Note that we do not obtain experimental data for TS-IDS on the NF-CSE-CIC-IDS2018 and NF-CSE-CIC-IDS2018-v2 datasets, because 
the provided code can not satisfy the requirements when performing data processing. 
So we directly excerpt the result for the NF-CSE-CIC-IDS2018-v2 dataset from the paper of TS-IDS \cite{H. Nguyen}, which is better than that of our method within 2\% in the two metrics.

\begin{figure*}[h!]
    \begin{minipage}{.5\textwidth}
            \centering
            \includegraphics[height = 0.15 \textheight]{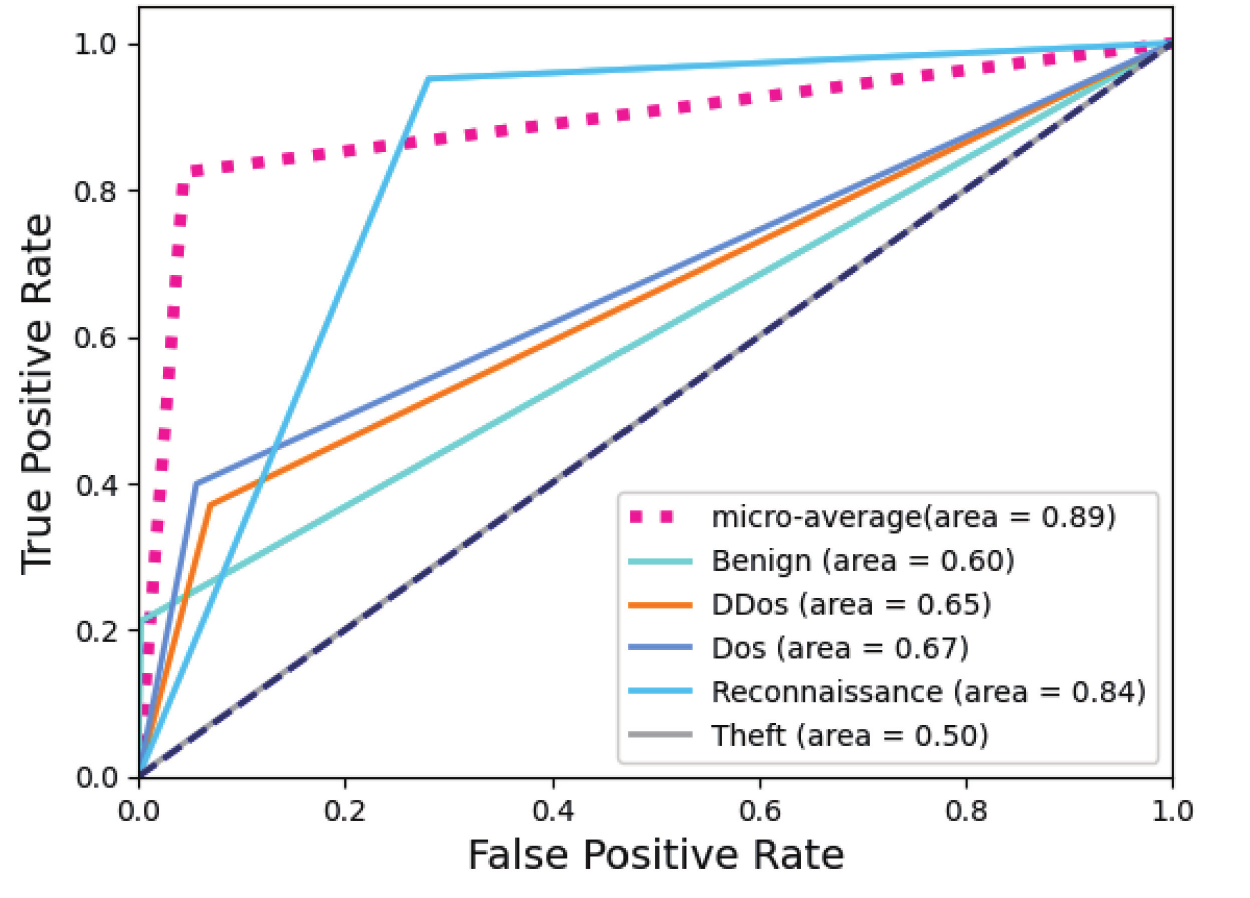}
            \caption{ROC curve of our method on NF-BoT-IoT.}
            \label{our roc bot}
        \end{minipage}%
        \begin{minipage}{.5\textwidth}
            \centering
            \includegraphics[height = 0.15 \textheight]{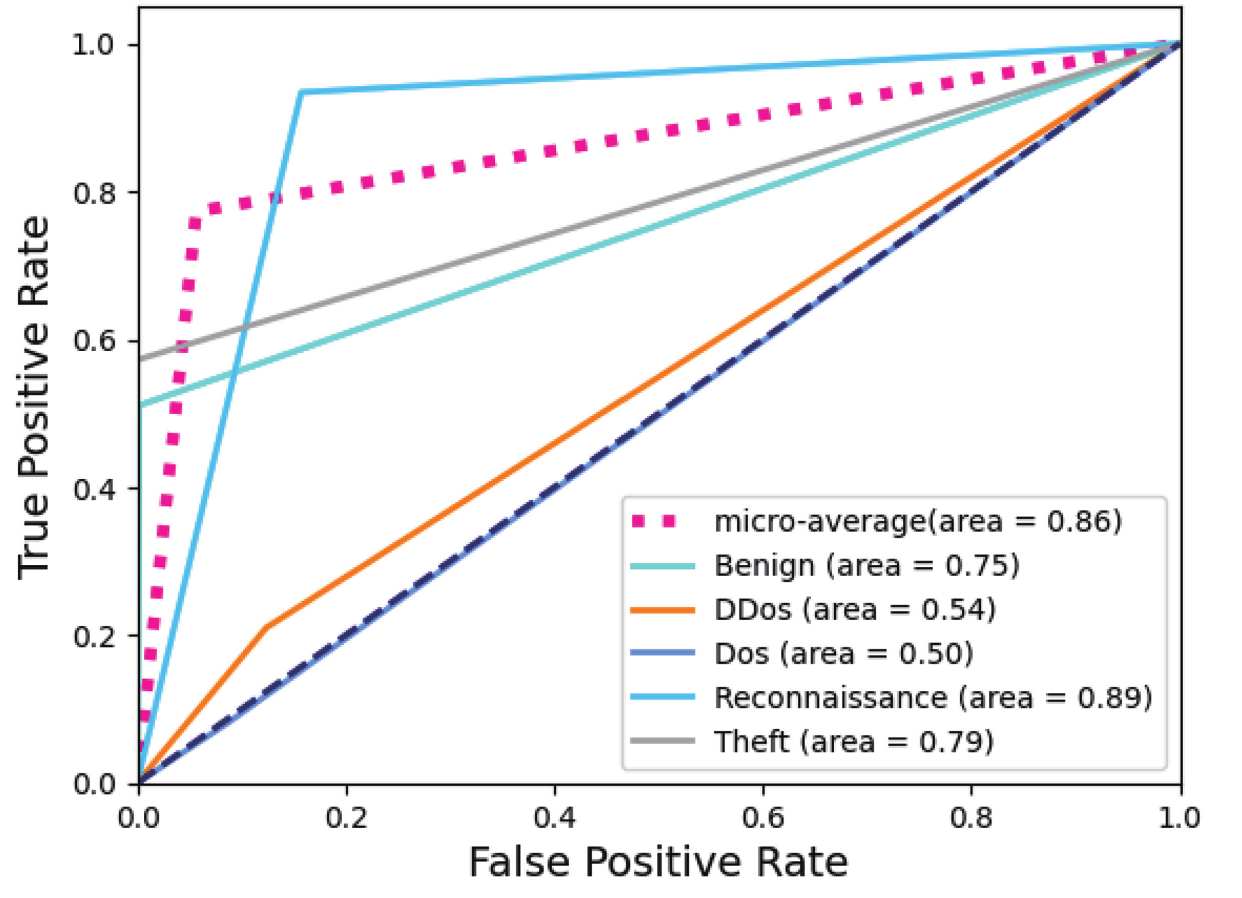}
            \caption{ROC curve of TS-IDS method on NF-BoT-IoT.}
            \label{ts roc bot}
        \end{minipage}%
    \end{figure*}

The Receiver Operating Characteristic (ROC) curve comprehensively shows the performance of the model under different thresholds, including the true positive rate and the false positive rate, and the ROC curve reflects the model performance stably even if the distribution of positive and negative samples in the testing set varies.
Due to the failure of completing the experiments of TS-IDS on the NF-CSE-CIC-IDS2018 and NF-CSE-CIC-IDS2018-v2 datasets, only the ROC curves of the NF-BoT-IoT and NF-BoT-IoT-v2 datasets are shown.
Fig. \ref{our roc bot} and Fig. \ref{ts roc bot} describe the ROC curves of our method and TS-IDS for the NF-BoT-IoT and  NF-BoT-IoT-v2 datasets, respectively.
The micro-average of the ROC curve by our method in the NF-BoT-IoT dataset is 89\%, that is better than 86\% of the TS-IDS method.  
Specifically, compare to the TS-IDS method, the areas of the ROC curves of our method for the DoS and DDoS samples are significant larger, while the areas for the Reconnaissance and Theft samples are smaller.

    \begin{figure*}[h!]
    \begin{minipage}{.5\textwidth}
        \centering
        \includegraphics[height = 0.15 \textheight]{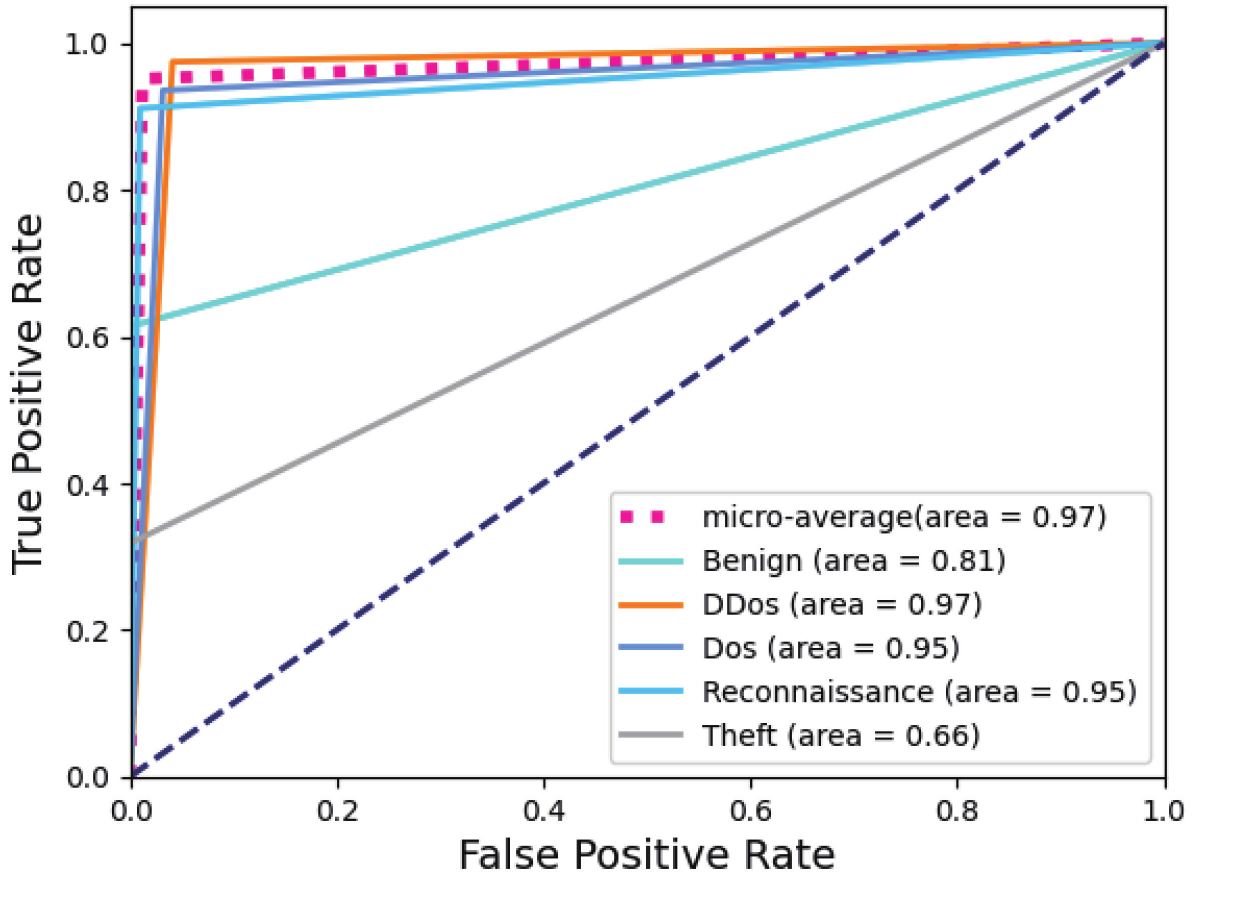}
        \caption{ROC curve of our method on NF-BoT-IoT-v2.}
        \label{our roc bot v2}
    \end{minipage}%
    \begin{minipage}{.5\textwidth}
        \centering
        \includegraphics[height = 0.15 \textheight]{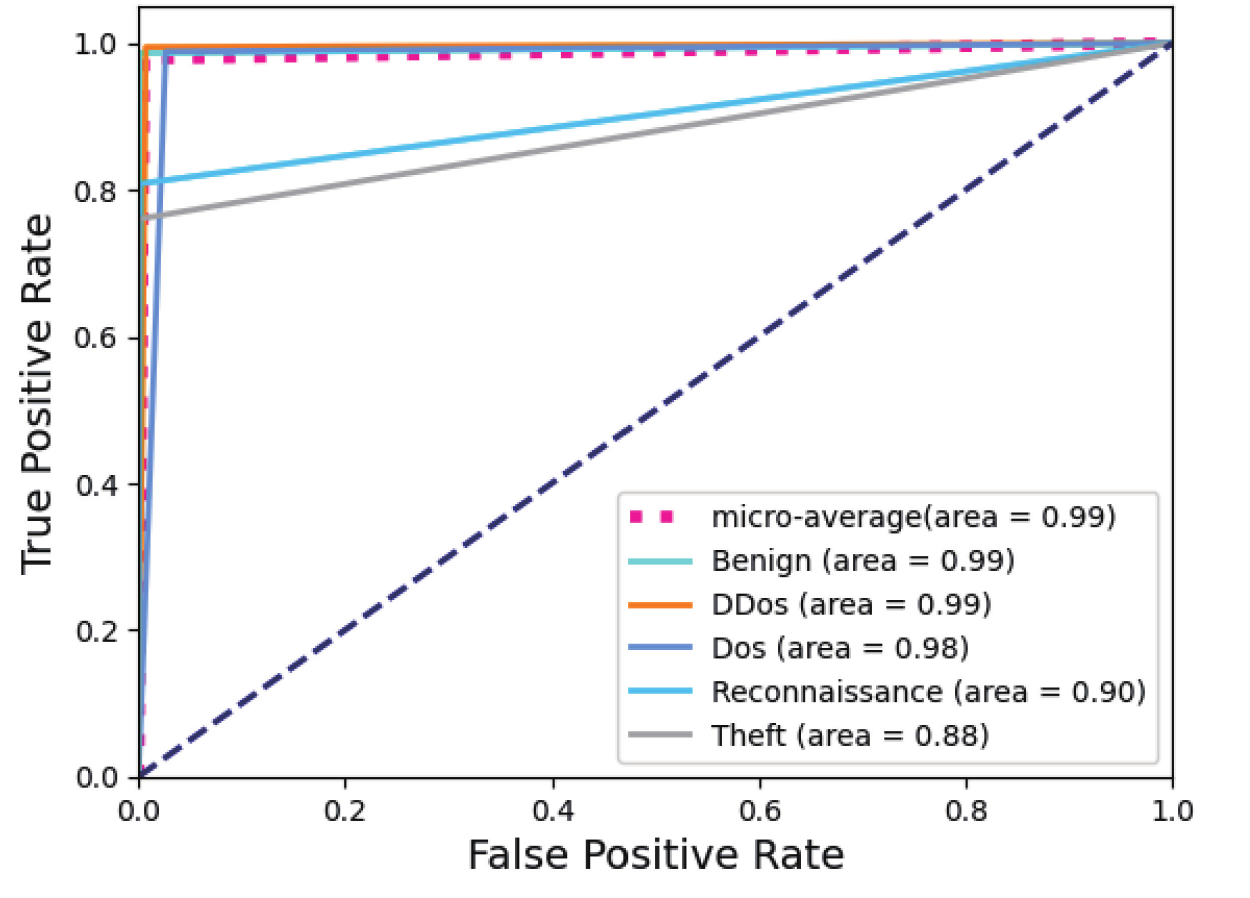}
        \caption{ROC curve of TS-IDS method on NF-BoT-IoT-v2.}
        \label{ts roc bot v2}
    \end{minipage}%
\end{figure*}

Fig. \ref{our roc bot v2} and Fig. \ref{ts roc bot v2} report the ROC curve results of the TS-IDS method and our method on the NF-BoT-IoT-v2 dataset, respectively. 
The micro-average of the ROC curve for our method is 97\%, which is less than 99\% achieved by TS-IDS. 
That implies that our method slightly underperforms the TS-IDS method on NF-BoT-IoT-v2.
In fact, the areas of the ROC curves for different types all match those of TS-IDS, except the Benign and Theft samples that only account for a small portion of the data.
It may be because our self-supervised method needs more corresponding data for training than that of supervised method.
    \begin{table}[h!]
        \footnotesize
        \centering
        \captionsetup{
            singlelinecheck=false, 
            justification=raggedright,
            width=0.327\textwidth
        }
        \caption{Comparison of the result of our method with TS-IDS method for multiclass classification.} 
        \label{Proposed Multi results}
        \begin{tabular}{cccc}
            \toprule[0.1em]
            \textbf{Model} & \textbf{Dataset} & \textbf{Recall} & \textbf{F1} \\
            \midrule[0.1em]
            TS-IDS & \multirow{2}{*}{NF-BoT-IoT} & 77.29\% & 78.20\%  \\
            Ours & & \bf{83.06\%} & \bf{82.70\%} \\
            \hline
            TS-IDS & \multirow{2}{*}{\shortstack{NF-BoT-IoT-v2}}& \bf{97.89\% }& \bf{97.84\% }\\
            Ours & & 95.16\% & 95.15\%\\
            \hline
            TS-IDS & \multirow{2}{*}{\shortstack{NF-CSE-CIC\\-IDS2018 }} & - & -  \\
            Ours & & 95.79\% & 94.79\% \\
            \hline
            TS-IDS & \multirow{2}{*}{\shortstack{NF-CSE-CIC\\-IDS2018-v2 }}& \bf{99.03\%} & \bf{98.08\%}  \\
            Ours & & 97.85\% & 97.43\% \\
            \bottomrule[0.1em]
        \end{tabular}
      \end{table}

\subsubsection{ Ablation Experiments }

To verify the effectiveness of the encoder NEGAT and self-supervised GNN framework NEGSC used in our method, we perform a series of ablation experiments, and list the detailed results of these experiments in Table \ref{ABLATION EXPERIMENTS}.
In the first experiment, to demonstrate the effectiveness of our proposed self-supervised framework NEGSC, we keep using the proposed NEGAT as encoder and replace NEGSC with original GSC \cite{Y. Han}. 
As can be seen from Table \ref{ABLATION EXPERIMENTS}, on the four datasets NF-BoT-IoT, NF-BoT-IoT-v2, NF-CSE-CIC-IDS2018 and NF-CSE-CIC-IDS2018-v2, our method (NEGAT + NEGSC) significantly outperform NEGAT+GSC in Rcall and F1 Metrics.
Compared to GSC, our proposed NEGSC is capable of handling both edge and node information, and thus is more suitable for NIDS where edge information plays a crucial role. 
In the second experiment, to verify the effectiveness of the proposed encoder NEGAT, we use the state-of-art method E-GraphSage \cite{W.W. Lo} to replace NEGAT as encoder and feed the graph embedding output of E-GraphSage to NEGSC. 
From Table \ref{ABLATION EXPERIMENTS}, it can be found that the value of Recall and F1 decrease after replacing the encoder by E-GraphSage on all four databases, which shows that our encoder outperforms E-GraphSAGE in network intrusion detection tasks. 
The reason is that the attention mechanism introduced by NEGAT can help our model better understand the local structure of graph, and more accurately utilize the different impacts of related traffic flows. 

\begin{table}[h]
    \footnotesize
    \centering
    \captionsetup{
        singlelinecheck=false, 
        justification=raggedright,
        width=0.441\textwidth
    }
    \caption{Results of ablation experiments.} 
    \label{ABLATION EXPERIMENTS}
    \begin{tabular}{lccc}
            \toprule[0.1em]
            \multicolumn{1}{c}{\textbf{Experimental Seting}} & \textbf{Dataset} & \textbf{Recall} & \textbf{F1} \\
            \midrule[0.1em]
            \textbf{NEGAT+NEGSC (ours)} & \makecell{NF-BoT-IoT} & \bf{83.06\%} & \bf{82.70\%} \\
            \textbf{NEGAT+GSC} & \makecell{NF-BoT-IoT} & 79.27\% & 72.14\% \\
            \textbf{\makecell{E-GraphSAGE+NEGSC}} & \makecell{NF-BoT-IoT} & 82.71\% & 82.19\% \\
            \hline
            \textbf{NEGAT+NEGSC (ours)} & \makecell{NF-BoT-IoT\\-v2} & \bf{95.16\%} & \bf{95.15\%} \\
            \textbf{NEGAT+GSC} & \makecell{NF-BoT-IoT\\-v2} & 90.21\% & 90.13\% \\
            \textbf{\makecell{E-GraphSAGE+NEGSC}} & \makecell{NF-BoT-IoT\\-v2} & 94.33\% & 94.27\% \\
            \hline
            \textbf{NEGAT+NEGSC (ours)} & \makecell{NF-CSE-CIC \\ -IDS2018} & \bf{95.79\%} & \bf{94.79\%} \\
            \textbf{NEGAT+GSC} & \makecell{NF-CSE-CIC \\ -IDS2018} & 85.03\% & 87.52\% \\
            \textbf{\makecell{E-GraphSAGE+NEGSC}} & \makecell{NF-CSE-CIC \\ -IDS2018} & 87.81\% & 83.93\% \\
            \hline
            \textbf{NEGAT+NEGSC (ours)} & \makecell{NF-CSE-CIC- \\ IDS2018-v2} & \bf{97.85\%} & \bf{97.43\%} \\
            \textbf{NEGAT+GSC} & \makecell{NF-CSE-CIC- \\ IDS2018-v2} & 88.36\% & 88.23\% \\
            \textbf{\makecell{E-GraphSAGE+NEGSC}} & \makecell{NF-CSE-CIC- \\ IDS2018-v2} & 97.42\% & 96.69\% \\
            \bottomrule[0.1em]
    \end{tabular}
\end{table}

\section{Conclusion}
\label{5}
This paper presented a self-supervised graph neural network for network intrusion detection systems, that is designed to efficiently and elaborately differentiate the normal network flows and the malicious flows with different attack types. 
To the best of our knowledge, this is the first GNN-based method for multiclass classification tasks in NIDS in an unsupervised manner. 
Extensive experimental evaluations on multiple netflow-based datasets demonstrate the preliminary generalization capability and application potential of our method in real-world traffic detection scenarios.
In addition, although here the proposed NEGSC is used for the edge-centered field NIDS, it is actually capable of simultaneously handling node and edge information, and thus may find applications in other scenarios.
Our future work for NIDS will focus on solving the problem of data imbalance in datasets and consider the timing dependency of network flows in real network environments.





\section*{Declaration of competing interest}

The authors declare that they have no known competing financial interests or personal relationships that could have appeared to influence the work reported in this paper.

\section*{Author Statements}

Conceptualization: R. Xu, G. Wu, W. Wang and A. He; Investigation: R. Xu, X. Gao and Z. Zhang; Writing - original draft: R. Xu, G. Wu and X. Gao; Methodology: R. Xu G. Wu and Z. Zhang; Software: R. Xu; Validation: G. Wu; Visualization: R. Xu and Z. Zhang; Resources: G. Wu, W. Wang and A. He; Funding acquisition: G. Wu, W. Wang and A. He; Writing - review and editing: G. Wu and X. Gao; Supervision: G. Wu. All authors have read and agreed to the version of the manuscript.

\section*{Acknowledgments}

This work is supported by the National Natural Science Foundation of China under Grant 62272486,
the Natural Science Foundation of Hunan Province, China, under Grant 2020JJ4949,  
the Key Project of Teaching Research of Hunan Province under Grant HNJG-2022-0124.

\bibliographystyle{plain}

\end{document}